\documentstyle[12pt]{article}
\newtheorem{definition}{Definition}[section]
\newtheorem{example}{Example}[section]

\newtheorem{lemma}{Lemma}[section]
\newtheorem{theorem}[lemma]{Theorem}

\newtheorem{algorithm}{Algorithm}[section]

\begin{document}

\title{\Large \bf Verifying Termination of General Logic Programs
with Concrete Queries}

\author{Yi-Dong Shen\thanks{Current corresponding address:
Department of Computing Science, 
University of Alberta, Edmonton, AB, Canada T6G 2H1.
Email: ydshen@cs.ualberta.ca}\\
{\small  Department of Computer Science,
Chongqing University, Chongqing 400044, China}\\
{\small Email: ydshen@cs.ualberta.ca}\\
Li-Yan Yuan  and Jia-Huai You\\
{\small  Department of Computing Science, University of
Alberta, Edmonton, Canada T6G 2H1}\\
{\small  Email: \{yuan, you\}@cs.ualberta.ca}}

\date{}

\maketitle

\begin{abstract}
We introduce a method of verifying termination of logic programs
with respect to concrete queries (instead of abstract query patterns). 
A necessary and sufficient condition is established 
and an algorithm for automatic verification is developed.
In contrast to existing query pattern-based approaches, our method has the following
features: (1) It applies to all general 
logic programs with non-floundering queries. (2) It is very easy to  
automate because it does not need to search for
a level mapping or a model, nor does it need to compute an interargument
relation based on additional mode or type information. (3) It bridges 
termination analysis with loop checking, the two problems
that have been studied separately in the past despite their close 
technical relation with each other.\\[.16in]  

\end{abstract}

\section{Introduction}
For a program in any computer language, in addition to having to be logically
correct, it should be terminating. Due to the recursive nature of
logic programming, however, a logic program may more likely be
non-terminating than a procedural program. Termination of
logic programs then becomes one of the most important topics in
logic programming research. Because the problem is extremely hard (undecidable
in general), it has been considered as a {\em never-ending story}; see
\cite{DD93,DDV99} for a comprehensive survey.

The goal of termination analysis is to establish a characterization
of termination of a logic program and design algorithms for automatic
verification. A lot of methods for termination analysis have been
proposed in the last decade (e.g., see
\cite{Apt1,Bezem92,BCF94,DV95,DDV99,Dev90,EBC99,KRKS98,Pl90a,Pl90b,UVG88,VD91}).
A majority of these existing methods are the {\em norm-} or {\em level 
mapping-based} approaches, which consist of inferring mode/type information, inferring
norms/level mappings, inferring models/interargument relations, and verifying
some well-founded conditions (constraints). For example,
Ullman and Van Gelder \cite{UVG88} and 
Pl\"{u}mer \cite{Pl90a,Pl90b} focused on 
establishing a decrease in term size of some recursive calls based on
interargument relations; Apt, Bezem and Pedreschi \cite{Apt1,Bezem92},
and Bossi, Cocco and Fabris \cite{BCF94}
provided characterizations of Prolog left-termination based on level
mappings/norms and models; Verschaetse \cite{V92}, Decorte, De Schreye
and Fabris \cite{DDF93}, and Martin, King and Soper \cite{MKS96} exploited inferring
norms/level mappings from mode and type information;  
De Schreye and Verschaetse \cite{DV95}, Brodsky and Sagiv \cite{BS91},
and Lindenstrauss and Sagiv \cite{LS97} discussed
automatic inference of interargument/size relations; 
De Schreye, Verschaetse and Bruynooghe \cite{DVB92}, and Mesnard \cite{M96}
addressed automatic verification of the well-founded constraints.
Very recently, Decorte, De Schreye and Vandecasteele \cite{DDV99}
presented an elegant unified termination analysis that integrates all
the above components to produce a set of constraints that, when solved, 
yields a termination proof.

It is easy to see that the above methods have among others the 
following features.

\begin{enumerate}
\item
They are compile-time approaches in the sense that they make termination
analysis only relying on some {\em static} information about the structure
(of the source code) of a logic program, such as 
modes/types, norms (i.e. term sizes of atoms of clauses)/level 
mappings, models/interargument relations, and the like. 

\item
They are suitable for termination analysis with respect to
(abstract) query patterns \cite{DVB92}. A query pattern defines a
class of concrete queries,\footnote{The difference between an abstract
query pattern and a concrete query is similar to that between
a class and an object in object-oriented programming languages.}
such as ground queries, bounded queries, well-moded queries, etc.

\end{enumerate}

Our observation shows that some {\em dynamic}
information about the structure of a concrete infinite SLDNF-derivation,
such as {\em repetition} of selected subgoals and clauses and {\em recursive
increase} in term size, plays a crucial role in characterizing the 
termination. Such dynamic features are hard to capture
unless we evaluate some related concrete
queries. This suggests that methods of extracting and utilizing 
dynamic features for termination analysis should be exploited.

Another observation comes from real programming practices. Consider the 
following situation: Given a logic program $P$ and a 
query pattern $Q$, applying a termination analysis yields a conclusion
that $P$ is not terminating w.r.t. $Q$. In most cases, this means that
there are a handful of concrete queries of the pattern $Q$ evaluating
which may lead to infinite SLDNF-derivations. In order to improve 
the program, users (programmers) most often want to figure out 
how the non-termination happens by posing a few typical concrete
queries and evaluating them step by step while determining which
derivations would most likely extend to infinite ones.  
Such a debugging process is both quite time consuming and tricky.
Doing it automatically is of great significance.
Obviously, the above mentioned termination analysis techniques cannot
help with such job. This suggests that methods of termination
analysis for concrete queries should be developed. 

The above two observations motivated the research of this paper.
In this paper, we introduce an effective method for termination analysis
w.r.t. concrete queries. The basic idea is as follows: First,
since non-termination is caused by an infinite (generalized) SLDNF-derivation, 
we directly make use of some essential structural characteristics of
an infinite derivation (such as variants, expanded variants, etc.)
to characterize the termination. Then, given a logic program
and a set of concrete queries, we evaluate these queries 
while dynamically collecting and
applying certain structural features to predict (based on the characterization)
if we are on the track to an infinite derivation. Such a process of query evaluation
is guaranteed to terminate by a necessary condition of an infinite derivation.
Finally, we provide the user with either an answer {\em Yes}, meaning that
the logic program is terminating w.r.t. the given set of queries, or
a finite (generalized) SLDNF-derivation that would most likely lead to
an infinite derivation. In the latter case, the user can improve the
program following the guidance of the informative derivation.

Although the termination problem is undecidable in general, our method
works effectively for a vast majority of general logic programs with 
non-floundering queries. In fact, the methodology used in this paper
is partly borrowed from loop checking $-$ another research topic 
in logic programming, which focuses
on detecting and eliminating infinite loops in SLD-trees 
(e.g., see \cite{BAK91,BDM92,MD96,MDB92,Sa93,shen97,shen001,VG87,VL89}). Therefore,
our work bridges termination analysis with loop checking, the two problems
which have been studied separately in the past despite their close 
technical relation with each other \cite{DD93}.

The plan of the paper is as follows. In Section 2, we introduce a notion
of a generalized SLDNF-tree, which is the basis of our method. Roughly
speaking, a generalized SLDNF-tree is a set of SLDNF-trees augmented 
with an ancestor-descendant relation on their subgoals. In Section 3, we prove
a necessary and sufficient condition for an infinite generalized
SLDNF-derivation. In Section 4.1, we formally define the notion of termination,
which is slightly different from that of De Schreye and Decorte \cite{DD93}.
In Section 4.2, we develop an algorithm for automatically verifying 
termination of a general logic program with concrete queries
and prove its properties. We will
use some representative logic programs to illustrate the effectiveness
of the algorithm. In Section 5, we mention some related work on termination 
analysis and on loop checking. We end in Section 6 with some concluding
remarks and further work.

\subsection{Preliminary} 

We present our notation and review some standard 
terminology of logic programs as described in \cite{Ld87}.

Variables begin with a capital letter, and predicate, function 
and constant symbols with a lower case letter. Let $A$ be an atom/function.
The size of $A$, denoted $|A|$, is the count of function symbols, 
variables and constants in $A$.
We use $rel(A)$ to refer to the predicate/function symbol of $A$, 
and use $A[i]$ to refer to the $i$-th argument of $A$, $A[i][j]$ 
to refer to the $j$-th argument of the $i$-th argument, and so on.
Let $S$ be a set or a list. We use $|S|$ to denote the number of
elements in $S$.

\begin{definition}
{\em
Let $A$ be an atom with the list $[X_1,...,X_m]$ of distinct variables.
By {\em variable renaming} on $A$ we mean to substitute the variables
of $A$ with another list $[Y_1,...,Y_m]$ of distinct variables.
Two atoms $A$ and $B$ are said to be {\em variants} if after variable 
renaming (on $A$ or $B$) they become the same.
}
\end{definition}
 
For instance, let $A=p(a,X,Y,X)$ and $B=p(a,Z,Y,Z)$. By 
substituting $[X,Y]$ for $[Z,Y]$, $B$ becomes the same as $A$, 
so $A$ and $B$ are variants. However,
$A$ and $C=p(a,Z,Y,W)$ are not variants because there is no variable
substitution that makes them the same.
Note that any atom $A$ is a variant of itself.

\begin{definition}
{\em
A (general) {\em logic program} is a finite set
of clauses of the form

$\qquad A\leftarrow L_1,..., L_n$

\noindent where $A$ is an atom and $L_i$s are literals. 
$A$ is called the {\em head} and $L_1,...,L_n$ is called the
{\em body} of the clause. If a general logic program has no clause with negative
literals in its body, it is called a {\em positive} program.
}
\end{definition}

\begin{definition}
{\em
A {\em goal} is a headless clause
$\leftarrow L_1,..., L_n$ where each literal $L_i$ is called a {\em subgoal}.
$L_1,..., L_n$ is called a (concrete) {\em query}.
When $n=0$, the ``$\leftarrow$'' symbol is omitted. 
}
\end{definition}

The initial goal, $G_0=\leftarrow L_1,..., L_n$, is called
a {\em top} goal. Without loss of generality, we shall assume throughout
the paper that a top goal consists only of one atom (i.e. $n=1$ and $L_1$
is a positive literal). 

\begin{definition}
{\em
A {\em control strategy} consists of two rules: one rule for selecting 
one goal from among a set of goals, and one rule for selecting
one subgoal from the selected goal. 
}
\end{definition}

The second rule in a control strategy is usually called 
a {\em selection} or {\em computation}
rule in the literature. Throughout the paper we use a fixed 
{\em depth-first, left-most} control strategy (as in Prolog). 
So the {\em selected} subgoal in each goal is the
left-most subgoal.

Trees are commonly used to represent 
the search space of a top-down proof procedure. For convenience,
a node in such a tree is represented by $N_i:G_i$ where $N_i$ is the name
of the node and $G_i$ is a goal labeling the node. Assume no two 
nodes have the same name. Therefore, we can refer to nodes by their names.

\section{Generalized SLDNF-Trees}

Non-termination of general logic programs results from infinite
derivations. In order to characterize infinite
derivations more precisely, in this section we extend the standard SLDNF-trees 
\cite{Ld87} to include some new features. 

To characterize an infinite derivation we need first
to define the ancestor-descendant relation on its selected subgoals.
Informally, $A$ is an ancestor subgoal of $B$ if the proof of $A$
needs (or in other words goes via) the proof of $B$. For example,
let $M:\leftarrow A,A_1,...,A_m$ be a node in an SLDNF-tree, and
$N:\leftarrow B_1,...,B_n,A_1,...,A_m$ be a child node of $M$ that
is generated by resolving $M$ on the subgoal $A$ with a clause 
$A\leftarrow B_1,...,B_n$. Then $A$ at $M$ is an ancestor subgoal
of all $B_i$s at $N$. However, such relationship does not
exist between $A$ at $M$ and any $A_j$ at $N$. It is easily seen
that all $B_i$s at $N$ inherit the ancestor subgoals of $A$ at $M$.

The ancestor-descendant relation can be explicitly expressed using
an {\em ancestor list} introduced in \cite{shen97}, 
which is a set of pairs $(Node, Atom)$
where $Node$ is the name of a node and $Atom$ is the selected subgoal at 
$Node$. The ancestor list of a subgoal $L_j$ is $AL_{L_j}=\{(N_1,A_1),...,(N_k,A_k)\}$,
showing that $A_1$ at node $N_1$, ..., and $A_k$ at node $N_k$ are
all ancestor subgoals of $L_j$. For 
instance, in the above example, if the ancestor list of the subgoal $A$
at node $M$ is $AL_A$ then the ancestor list of each $B_i$ at node $N$
is $AL_{B_i}=\{(M,A)\}\cup AL_A$.

Augmenting SLDNF-trees with ancestor lists leads to the 
following definition of SLDNF$^*$-trees.

\begin{definition}[SLDNF$^*$-trees]
\label{tree}
{\em
Let $P$ be a general logic program, $G_0= \leftarrow A_0$ a top goal, 
and $R$ a depth-first, left-most control strategy.
The {\em SLDNF$^*$-tree} $T_{G_0}$
for $P \cup \{G_0\}$ via $R$ is defined as follows.
\begin{enumerate}
\item
\label{tree1}
The root node is $N_0:G_0$ with the ancestor list $AL_{A_0}=\{\}$
for $A_0$.

\item
\label{tree2}
Let $N_i: \leftarrow L_1,...,L_m$
be a node in the tree selected by $R$. If $m=0$ then
$N_i$ is a {\em success} leaf, marked by $\Box_t$.
Otherwise, we distinguish between the following two cases.

\begin{enumerate}
\item
\label{tree21}
If $L_1$ is a positive literal, then for each clause
$B \leftarrow B_1,...,B_n$ such that $L_1$ and $B$ 
are unifiable, $N_i$ has a child node

$\qquad N_s: \leftarrow (B_1,...,B_n,L_2,...,L_m)\theta$

\noindent where $\theta$ is an mgu (i.e. most general unifier) of $L_1$
and $B$, the ancestor list for each $B_k\theta$ is 
$AL_{B_k\theta}=\{(N_i,L_1)\}\cup AL_{L_1}$,
and the ancestor list for each $L_k\theta$ is $AL_{L_k\theta}=AL_{L_k}$.
If there exists no clause whose head can unify with $L_1$ then
$N_i$ has a single child  node $-$ a {\em failure} leaf, marked by $\Box_f$. 
 
\item
\label{tree22}
If $L_1=\neg A$ is a ground negative literal, then build a partial
SLDNF$^*$-tree $T_{\leftarrow A}$ for $P\cup \{\leftarrow A\}$
via $R$ where $A$ inherits the ancestor list of $L_1$, until
the first success leaf is generated. If $T_{\leftarrow A}$ has
a success leaf then $N_i$ has a single child node 
$-$ a failure leaf, $\Box_f$. Otherwise, if all branches of
$T_{\leftarrow A}$ end with a failure leaf then 
$N_i$ has a single child node 

$\qquad N_s:\leftarrow L_2,...,L_m$

\noindent where all $L_k$ inherit the ancestor lists of $L_k$ at node $N_i$.

\end{enumerate}

\end{enumerate}
}
\end{definition}

Note that in this paper we do not discuss floundering $-$ a situation where
a non-ground negative subgoal is selected by $R$
(see \cite{chan88, dra95, LAC99, Prz89-3} for discussion on such topic). 
In contrast to SLDNF-trees, an SLDNF$^*$-tree has the following two
new features.
\begin{enumerate}
\item
An ancestor list $AL_{L_j}$ is attached to each subgoal $L_j$. In particular, subgoals
of a subsidiary SLTNF$^*$-tree $T_{\leftarrow A}$ built for solving a subgoal
$L_1=\neg A$ inherit the ancestor list of $L_1$ (see item \ref{tree22}).
This is especially useful in identifying infinite derivations across
SLTNF$^*$-trees (see Example \ref{eg1}). Note that a negative subgoal
will never be an ancestor subgoal.

\item
To handle a ground negative subgoal $L_1=\neg A$, only a partial
subsidiary SLTNF$^*$-tree $T_{\leftarrow A}$ is generated by stopping
at the first success leaf (see item \ref{tree22}). 
The reason for this is that it is totally
unnecessary to exhaust the remaining branches of $T_{\leftarrow A}$ 
because they would have no new answer for $A$. This can not only improve
the efficiency of query evaluation, but also avoid some possible
infinite derivations (see Example \ref{eg2}). In fact, Prolog  
achieves this by using cuts to skip the remaining branches
of $T_{\leftarrow A}$ (e.g. see SICStus Prolog \cite{SICSTUS}).
  
\end{enumerate}

For convenience, we use dotted edges $``\cdot\cdot\cdot\triangleright''$
to connect parent and child SLDNF$^*$-trees, 
so that infinite derivations across SLDNF$^*$-trees can be clearly identified.
Moreover, we refer to $T_{G_0}$, the {\em top} SLDNF$^*$-tree, 
along with all its descendant SLDNF$^*$-trees
as a {\em generalized SLDNF-tree} for $P\cup \{G_0\}$, denoted $GT_{G_0}$.
Therefore, a path of a generalized SLDNF-tree may come 
across several SLDNF$^*$-trees through dotted edges. Any such a path
starting at the root node $N_0:G_0$ is called a {\em generalized 
SLDNF-derivation}. A generalized SLDNF-derivation is 
{\em successful} (resp. {\em failed}) if it ends at a success leaf
(resp. at a failure leaf).

Thus, there may occur two types of edges in a generalized SLDNF-tree,
``$\stackrel{C}{\longrightarrow}$''
and $``\cdot\cdot\cdot\triangleright''$. For convenience,
we use $``\Rightarrow''$ to refer to either of them.
We also use $N_i:G_i\stackrel{C_1}{\longmapsto}...\stackrel{C_m}{\longmapsto}N_k:G_k$
to represent a segment of a generalized SLDNF-derivation, which
generates $N_k:G_k$ from $N_i:G_i$ by applying
the set of clauses $\{C_1,...C_m\}$.
Moreover, for any node $N_i:G_i$ we use
$L_i^1$ to refer to the selected (i.e. left-most) subgoal in $G_i$.

\begin{example}
\label{eg1}
{\em
Let $P_1$ be a general logic program and $G_0$ a top goal, given by
\begin{tabbing}
$\qquad$ \= $P_1:$ $\quad$ \= $p(X)\leftarrow \neg p(f(X))$. \`$C_{p_1}$\\
\>          $G_0:$ \> $\leftarrow p(a).$
\end{tabbing}
The generalized SLDNF-tree $GT_{\leftarrow p(a)}$ for $P_1\cup \{G_0\}$ is shown in
Figure \ref{fig1}, where $\infty$ represents an infinite extension.
We see that $GT_{\leftarrow p(a)}$ consists of one infinite 
generalized SLDNF-derivation.

\begin{figure}[htb]
\centering
\setlength{\unitlength}{3947sp}%
\begingroup\makeatletter\ifx\SetFigFont\undefined%
\gdef\SetFigFont#1#2#3#4#5{%
  \reset@font\fontsize{#1}{#2pt}%
  \fontfamily{#3}\fontseries{#4}\fontshape{#5}%
  \selectfont}%
\fi\endgroup%
\begin{picture}(3975,1716)(676,-1006)
\thinlines
\put(2251, 89){\vector( 1, 0){0}}
\multiput(1876, 89)(75.00000,0.00000){5}{\makebox(1.6667,11.6667){\SetFigFont{5}{6}{\rmdefault}{\mddefault}{\updefault}.}}
\put(4051,-436){\vector( 1, 0){0}}
\multiput(3676,-436)(75.00000,0.00000){5}{\makebox(1.6667,11.6667){\SetFigFont{5}{6}{\rmdefault}{\mddefault}{\updefault}.}}
\put(4501,-961){\makebox(0,0)[lb]{\smash{\SetFigFont{10}{12.0}{\rmdefault}{\mddefault}{\updefault}$\infty$}}}
\put(1201,539){\vector( 0,-1){300}}
\put(2776, 14){\vector( 0,-1){300}}
\put(4576,-511){\vector( 0,-1){300}}
\put(1276,389){\makebox(0,0)[lb]{\smash{\SetFigFont{8}{9.6}{\rmdefault}{\mddefault}{\updefault}$C_{p_1}$}}}
\put(751,614){\makebox(0,0)[lb]{\smash{\SetFigFont{9}{10.8}{\rmdefault}{\mddefault}{\updefault}$N_0$:  $\leftarrow p(a)$}}}
\put(2851,-136){\makebox(0,0)[lb]{\smash{\SetFigFont{8}{9.6}{\rmdefault}{\mddefault}{\updefault}$C_{p_1}$}}}
\put(2326, 89){\makebox(0,0)[lb]{\smash{\SetFigFont{9}{10.8}{\rmdefault}{\mddefault}{\updefault}$N_2$:  $\leftarrow p(f(a))$}}}
\put(2251,-436){\makebox(0,0)[lb]{\smash{\SetFigFont{9}{10.8}{\rmdefault}{\mddefault}{\updefault}$N_3$:  $\leftarrow \neg p(f(f(a)))$ }}}
\put(676, 89){\makebox(0,0)[lb]{\smash{\SetFigFont{9}{10.8}{\rmdefault}{\mddefault}{\updefault}$N_1$:  $\leftarrow \neg p(f(a))$ }}}
\put(4126,-436){\makebox(0,0)[lb]{\smash{\SetFigFont{9}{10.8}{\rmdefault}{\mddefault}{\updefault}$N_4$:  $\leftarrow p(f(f(a)))$}}}
\put(4651,-661){\makebox(0,0)[lb]{\smash{\SetFigFont{8}{9.6}{\rmdefault}{\mddefault}{\updefault}$C_{p_1}$}}}
\end{picture}

\caption{A generalized SLDNF-tree $GT_{\leftarrow p(a)}$.}\label{fig1}
\end{figure}
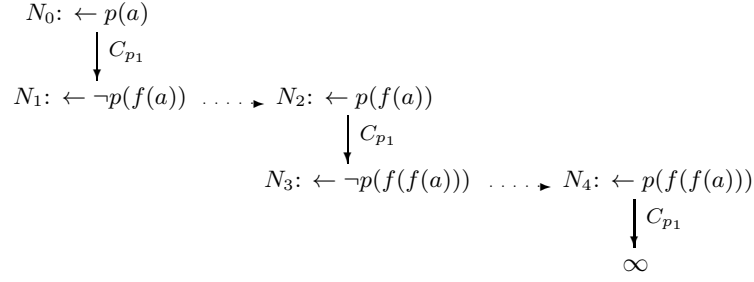
}
\end{example}

\begin{example}
\label{eg2}
{\em
Consider the following general logic program and top goal.
\begin{tabbing}
$\qquad$ \= $P_2:$ $\quad$ \= $p\leftarrow \neg q$. \`$C_{p_1}$\\
\>\> $q$.       \`$C_{q_1}$ \\
\>\> $q \leftarrow q$.            \`$C_{q_2}$ \\
\>          $G_0:$ \> $\leftarrow p.$
\end{tabbing}
The generalized SLDNF-tree $GT_{\leftarrow p}$ for $P_2\cup \{G_0\}$ is depicted in
Figure \ref{fig2} (a). For the purpose of comparison, 
the SLDNF-trees for $P_2\cup \{\leftarrow p\}$ 
are shown in Figure \ref{fig2} (b). Note that Figure \ref{fig2} (a) is finite, 
whereas Figure \ref{fig2} (b) is not.

\begin{figure}[h]
\centering
\setlength{\unitlength}{3947sp}%
\begingroup\makeatletter\ifx\SetFigFont\undefined%
\gdef\SetFigFont#1#2#3#4#5{%
  \reset@font\fontsize{#1}{#2pt}%
  \fontfamily{#3}\fontseries{#4}\fontshape{#5}%
  \selectfont}%
\fi\endgroup%
\begin{picture}(4512,2046)(826,-1336)
\thinlines
\put(1201,-361){\vector( 0,-1){0}}
\multiput(1201, 14)(0.00000,-75.00000){5}{\makebox(1.6667,11.6667){\SetFigFont{5}{6}{\rmdefault}{\mddefault}{\updefault}.}}
\put(4726,539){\vector(-2,-3){150}}
\put(4876,539){\vector( 2,-3){150}}
\put(5026, 14){\vector(-2,-3){150}}
\put(5176, 14){\vector( 2,-3){150}}
\put(1126,-1336){\makebox(0,0)[lb]{\smash{\SetFigFont{10}{12.0}{\rmdefault}{\mddefault}{\updefault}(a)}}}
\put(3901,-1336){\makebox(0,0)[lb]{\smash{\SetFigFont{10}{12.0}{\rmdefault}{\mddefault}{\updefault}(b)}}}
\put(5326,-436){\makebox(0,0)[lb]{\smash{\SetFigFont{10}{12.0}{\rmdefault}{\mddefault}{\updefault}$\infty$}}}
\put(1201,-586){\vector( 0,-1){300}}
\put(3376,539){\vector( 0,-1){300}}
\put(1201,539){\vector( 0,-1){300}}
\put(1276,389){\makebox(0,0)[lb]{\smash{\SetFigFont{8}{9.6}{\rmdefault}{\mddefault}{\updefault}$C_{p_1}$}}}
\put(826,614){\makebox(0,0)[lb]{\smash{\SetFigFont{9}{10.8}{\rmdefault}{\mddefault}{\updefault}$N_0$:  $\leftarrow p$}}}
\put(826, 89){\makebox(0,0)[lb]{\smash{\SetFigFont{9}{10.8}{\rmdefault}{\mddefault}{\updefault}$N_1$:  $\leftarrow \neg q$ }}}
\put(826,-511){\makebox(0,0)[lb]{\smash{\SetFigFont{9}{10.8}{\rmdefault}{\mddefault}{\updefault}$N_2$:  $\leftarrow q$}}}
\put(901,-1036){\makebox(0,0)[lb]{\smash{\SetFigFont{9}{10.8}{\rmdefault}{\mddefault}{\updefault}$N_3$:  $\Box_t$}}}
\put(1276,-736){\makebox(0,0)[lb]{\smash{\SetFigFont{8}{9.6}{\rmdefault}{\mddefault}{\updefault}$C_{q_1}$}}}
\put(3451,389){\makebox(0,0)[lb]{\smash{\SetFigFont{8}{9.6}{\rmdefault}{\mddefault}{\updefault}$C_{p_1}$}}}
\put(3001,614){\makebox(0,0)[lb]{\smash{\SetFigFont{9}{10.8}{\rmdefault}{\mddefault}{\updefault}$N_0$:  $\leftarrow p$}}}
\put(3001, 89){\makebox(0,0)[lb]{\smash{\SetFigFont{9}{10.8}{\rmdefault}{\mddefault}{\updefault}$N_1$:  $\leftarrow \neg q$ }}}
\put(4501,614){\makebox(0,0)[lb]{\smash{\SetFigFont{9}{10.8}{\rmdefault}{\mddefault}{\updefault}$N_2$:  $\leftarrow q$}}}
\put(4201, 89){\makebox(0,0)[lb]{\smash{\SetFigFont{9}{10.8}{\rmdefault}{\mddefault}{\updefault}$N_3$:  $\Box_t$}}}
\put(4351,389){\makebox(0,0)[lb]{\smash{\SetFigFont{8}{9.6}{\rmdefault}{\mddefault}{\updefault}$C_{q_1}$}}}
\put(4801, 89){\makebox(0,0)[lb]{\smash{\SetFigFont{9}{10.8}{\rmdefault}{\mddefault}{\updefault}$N_4$:  $\leftarrow q$}}}
\put(5026,389){\makebox(0,0)[lb]{\smash{\SetFigFont{8}{9.6}{\rmdefault}{\mddefault}{\updefault}$C_{q_2}$}}}
\put(4501,-436){\makebox(0,0)[lb]{\smash{\SetFigFont{9}{10.8}{\rmdefault}{\mddefault}{\updefault}$N_5$:  $\Box_t$}}}
\put(4651,-136){\makebox(0,0)[lb]{\smash{\SetFigFont{8}{9.6}{\rmdefault}{\mddefault}{\updefault}$C_{q_1}$}}}
\put(5326,-136){\makebox(0,0)[lb]{\smash{\SetFigFont{8}{9.6}{\rmdefault}{\mddefault}{\updefault}$C_{q_2}$}}}
\end{picture}

\caption{A generalized SLDNF-tree $GT_{\leftarrow p}$ (a)
and its two corresponding SLDNF-trees (b).}\label{fig2}
\end{figure}
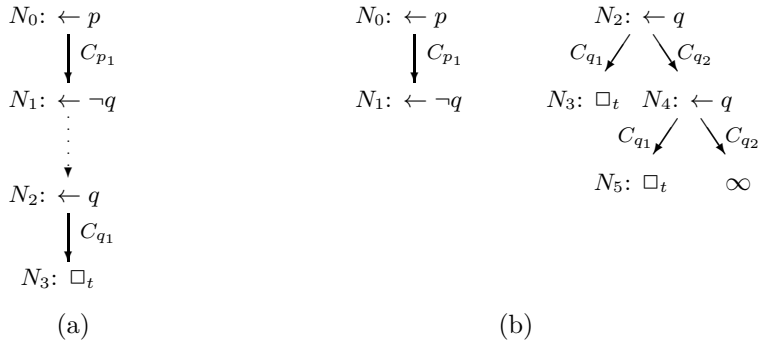
}
\end{example}

We now formally define the ancestor-descendant relation.

\begin{definition}
\label{asubgoal}
{\em
Let $N_i:G_i$ and $N_k:G_k$ be two nodes in a 
generalized SLDNF-derivation, and
$A$ and $B$ be the selected subgoals in $G_i$ and $G_k$,
respectively. We say that $A$ is an {\em ancestor subgoal} of $B$,
denoted $A\prec_{ANC}B$, if $A$ is in the ancestor list $AL_B$ of $B$.
When $A$ is an ancestor subgoal of $B$, we refer to $B$ as a {\em descendant
subgoal} of $A$, $N_i$ as an {\em ancestor node} of $N_k$, and $N_k$ as 
a {\em descendant node} of $N_i$. 
}
\end{definition}

\section{Characterizing an Infinite Generalized SLDNF-Derivation}

In this section we establish a necessary and sufficient condition for
an infinite generalized SLDNF-derivation.

In \cite{shen001}, a concept of expanded variants is introduced,
which captures some key structural characteristics of certain
subgoals in an infinite SLD-derivation. We observe that it applies
to general logic programs as well. That is, infinite generalized 
SLDNF-derivations can be characterized based on expanded variants.

\begin{definition}
\label{gvar}
{\em
Let $A$ and $A'$ be two atoms or functions.
$A'$ is said to be an {\em expanded variant} of $A$,
denoted $A'\sqsupseteq_{EV}A$,
if after variable renaming on $A'$
it becomes $B$ that is the same as $A$ except that there may be
some terms at certain positions in $A$
each $A[i]...[k]$ of which grows in $B$ into a function 
$B[i]...[k]=f(...,A[i]...[k],...)$. Such terms like
$A[i]...[k]$ in $A$ are then called {\em growing terms}
w.r.t. $A'$.
}
\end{definition}

As an illustration, let $A=p(X,g(X))$ and $A'=p(Y, g(h(Y)))$. By renaming $Y$ with
$X$, $A'$ becomes $B=p(X,g(h(X)))$, which is the same 
as $A$ except that $B[2][1]=h(A[2][1])$. Therefore, $A'$ is an expanded
variant of $A$ with a growing term $A[2][1]$. Here are a few more examples:
$p(X,Y)\sqsupseteq_{EV} p(Z,W)$, $p(f(a))\sqsupseteq_{EV} p(a)$, 
$p(g(a),f(g(h(X))))\sqsupseteq_{EV} p(a,f(h(Y)))$,
and $p([X_1,X_2,X_3])\sqsupseteq_{EV} p([X_1,X_4])$
(note that $[X_1,X_2,X_3]=[X_1|[X_2,X_3]]$).

It is immediate from Definition \ref{gvar} that variants are expanded 
variants with the same size.

\begin{theorem}
\label{th2}
Let $D$ be an infinite generalized SLDNF-derivation 
without infinitely large subgoals. Then there are infinitely many goals
$G_{g_1}, G_{g_2}, ...$ in $D$ such that for any $j \geq 1$,
$L_{g_j}^1\prec_{ANC}L_{g_{j+1}}^1$ and $L_{g_j}^1$ and $L_{g_{j+1}}^1$ 
are variants.
\end{theorem}

\noindent {\bf Proof.}
Let $D$  be of the form
\begin{tabbing}
$\qquad$ $N_0:G_0\Rightarrow N_1:G_1 \Rightarrow ... \Rightarrow
N_i:G_i\Rightarrow N_{i+1}:G_{i+1} \Rightarrow ...$
\end{tabbing}
For each derivation step 
$N_i:G_i \stackrel{C}{\longrightarrow} N_{i+1}:G_{i+1}$, where 
$L_i^1$ is a positive subgoal and $C=A\leftarrow B_1,...B_n$
such that $A\theta = L_i^1\theta$ under an mgu $\theta$, we do the following: 
\begin{enumerate}
\item
If $n=0$, which means $L_i^1$ is proved at this step,
mark node $N_i$ with \#.

\item
Otherwise, the proof of $L_i^1$ needs the proof of $B_j\theta$ $(j=1,...,n)$.
If all descendant nodes of $N_i$ in $D$ have been marked with \#, which means that 
all $B_j\theta$ have been proved at some steps in $D$, mark node $N_i$
with \#.
\end{enumerate}

Note that the root node $N_0$ will never be marked by \#, for otherwise
$G_0$ would have been proved and $D$ should have ended at a success leaf. After the 
above marking process, let $D$ become
\begin{tabbing}
$\qquad$ $N_0:G_0\Rightarrow ... \Rightarrow N_{i_1}:G_{i_1} \Rightarrow ... \Rightarrow
N_{i_2}:G_{i_2}\Rightarrow ... \Rightarrow  N_{i_k}:G_{i_k} \Rightarrow ...$
\end{tabbing}
where all nodes except $N_0,N_{i_1},N_{i_2},...,N_{i_k},...$ are marked with
\#. Since we use the depth-first, left-most control strategy, for any 
$j\geq 0$ the proof of $L_{i_j}^1$ needs the proof of $L_{i_{j+1}}^1$ (let
$i_0=0$), for otherwise $N_{i_j}$ would have been marked with \#. That is,
$L_{i_j}^1$ is an ancestor subgoal of $L_{i_{j+1}}^1$. Moreover, $D$ must contain 
an infinite number of such nodes because if $N_{i_k}:G_{i_k}$ was the last one,
which means that all nodes after $N_{i_k}$ were marked with \#, then $L_{i_k}^1$
would be proved, so that $N_{i_k}$ should be marked with \#, a contradiction.

The above proof shows that $D$ has an infinite number of selected subgoals 
$L_{i_1}^1,L_{i_2}^1,...$ such that 
$L_{i_j}^1 \prec_{ANC} L_{i_{j+1}}^1$ $(j\geq 1)$.
Since all subgoals in $D$ are bounded in size, and any general logic program 
has only a finite number of clauses and predicate, function and constant
symbols, there must be an infinite number of subgoals 
$L_{g_1}^1,L_{g_2}^1,...$ among the
$L_{i_j}^1$s that are variants. This concludes the proof.

\begin{theorem}
\label{th3}
Let $D$ be an infinite  generalized SLDNF-derivation 
with infinitely large subgoals. Then there are infinitely many goals
$G_{g_1}, G_{g_2}, ...$ in $D$ such that for any $j \geq 1$,
$L_{g_j}^1\prec_{ANC}L_{g_{j+1}}^1$ and $L_{g_{j+1}}^1\sqsupseteq_{EV}L_{g_j}^1$ 
with $|L_{g_{j+1}}^1|>|L_{g_j}^1|$.
\end{theorem}

The following lemma is required to prove this theorem.

\begin{lemma}
\label{lemm31}
Let $S=\{C_1, ..., C_n\}$ be a finite set of clauses. 
Let $D$ be an infinite  generalized SLDNF-derivation
of the form
\begin{tabbing}
$\qquad$ $N_0:G_0\Rightarrow ... N_{i_1}:G_{i_1} 
\stackrel{C_{1_1}}{\longmapsto} ... \stackrel{C_{1_{n_1}}}{\longmapsto}
N_{i_2}:G_{i_2} 
\stackrel{C_{2_1}}{\longmapsto} ... \stackrel{C_{2_{n_2}}}{\longmapsto}
N_{i_3}:G_{i_3} 
\stackrel{C_{3_1}}{\longmapsto} ...$
\end{tabbing}
where for any $j \geq 1$, $\{C_{j_1}, ..., C_{j_{n_j}}\}=S$,
$L_{i_j}^1\prec_{ANC}L_{i_{j+1}}^1$,
and $L_{i_j}^1$ is a variant of $L_{i_{j+1}}^1$ 
except for a few terms (at least one) at certain positions in $L_{i_{j+1}}^1$ that
increase in size w.r.t. $L_{i_j}^1$. Then there are an infinite sequence of subgoals
$L_{g_1}^1,L_{g_2}^1...$ among the $L_{i_j}^1$s 
such that for any $k \geq 1$,
$L_{g_{k+1}}^1\sqsupseteq_{EV}L_{g_k}^1$ 
with $|L_{g_{k+1}}^1|>|L_{g_k}^1|$.
\end{lemma}

\noindent {\bf Proof.}
Since $L_{i_{j+1}}^1$ being an expanded variant of $L_{i_j}^1$ 
with $|L_{i_{j+1}}^1|>|L_{i_j}^1|$ is determined
by those arguments of $L_{i_j}^1$ whose size increases
w.r.t. $L_{i_{j+1}}^1$, for simplicity of presentation we
ignore the remaining arguments of $L_{i_j}^1$ that are variants
of the corresponding ones in $L_{i_{j+1}}^1$.
Since the $L_{i_j}^1$s are generated by repeatedly applying
the same set $S$ of clauses, the increase in their term size
must be made in a fixed, regular way (assume that our programs contain no 
built-in's such as $assert(.)$ and $retract(.)$). 
In order to facilitate the analysis of such regular 
increase in term size, with no loss in generality,
let each $L_{i_j}^1$ be of the form
$p([T_1^j,...,T_{m_j}^j])$, which contains a single argument that is a list,
such that the list of $L_{i_{j+1}}^1$ is derived from the list $L=[T_1^j,...,T_{m_j}^j]$ of
$L_{i_j}^1$ via a DELETE-ADD-SHUFFLE process which consists of
{\em deleting} the first $0\leq n^- \leq m_j$ elements from $L$,
then {\em adding} $n^+\geq n^-$ elements to the front of the remaining 
part of $L$, and finally {\em shuffling} the list of elements obtained 
(in a fixed way).\footnote{See Example \ref{eg6-0} for an illustration.}

Let $\vec{X}$ represent a sequence of elements, such as $X_1,X_2,...,X_m$.
We distinguish the following two cases based on the ADD operation.

\begin{enumerate}
\item
\label{lem1-item1}
The $n^+$ added elements are independent of the list $L$ of elements
of $L_{i_j}^1$. Since the same set $S$ of clauses
is applied, the set $E_1$ of elements added to 
$L_{i_2}^1$ from $L_{i_1}^1$ must be the same as or a variant of
the set added to $L_{i_3}^1$ from $L_{i_2}^1$ that 
must be the same as or a variant of
the set added to $L_{i_4}^1$ from $L_{i_3}^1$, and so on. Let 
$E_2\subseteq \{T_1^1,...,T_{m_1}^1\}$ be such that each $T\in E_2$
occurs in the derivation $D$ infinite times. That is, no $T\in E_2$
will be removed by the DELETE operation. (But all $T\in \{T_1^1,...,T_{m_1}^1\}-E_2$
will be deleted at some derivation steps in $D$ by the DELETE operation.)
Then there must be an infinite
sequence $L_{f_1}^1,L_{f_2}^1, ...$ among the $L_{i_j}^1$s such that
for any $k \geq 1$, all elements in $L_{f_k}^1$ come from
$E_1\cup E_2$ (or its variant).
Since $E_1\cup E_2$ contains only a finite number of elements,
no matter what shuffling approach is used, there must be 
an infinite sequence 
$L_{g_1}^1,L_{g_2}^1...$ among the $L_{f_j}^1$s
such that for any $k \geq 1$, let $L_{g_k}^1=p([S_1,...,S_n])$,
then after variable renaming $L_{g_{k+1}}^1$ will become like 
$p([\vec{S_1},...,\vec{S_n}])$ such that each $S_l$
is in $\vec{S_l}$. Obviously, these
$S_l$s in $L_{g_k}^1$ are growing terms w.r.t. $L_{g_{k+1}}^1$. That is,
$L_{g_{k+1}}^1\sqsupseteq_{EV}L_{g_k}^1$ with $|L_{g_{k+1}}^1|>|L_{g_k}^1|$. 

\item
Some of the $n^+$ added elements depend on the list of $L_{i_j}^1$.
For simplicity of presentation and without loss of generality assume that 
from $L_{i_1}^1$ to $L_{i_2}^1$ only one added element, say of the form
$f(A_1^1,g(A_2^1))$, depends on the list $[T_1^1,...,T_{m_1}^1]$ of  
$L_{i_1}^1$. Let $E_1$ be the set of elements added to $L_{i_2}^1$ that are independent
of the list $[T_1^1,...,T_{m_1}^1]$. That is, the set of added elements
from $L_{i_1}^1$ to $L_{i_2}^1$ is
$E_1\cup \{f(A_1^1,g(A_2^1))\}$. Let $E=E_1\cup \{T_1^1,...,T_{m_1}^1\}$.
Then $E$ can be considered to be the domain of the two arguments 
$A_1^1$, $A_2^1$ in $f(A_1^1,g(A_2^1))$, i.e. $A_1^1, A_2^1\in E$.
Since all the $L_{i_j}^1$s in the derivation $D$ are generated recursively
by applying the same set $S$ of clauses, for any $j\geq 1$ 
from $L_{i_j}^1$ to $L_{i_{j+1}}^1$ the set of added elements should be
$E_1\cup \{f(A_1^j,g(A_2^j))\}$ where $A_1^j, A_2^j\in E\cup 
\{f(A_1^k,g(A_2^k))|k<j\}$. It is easy to see that any element of 
$L_{i_j}^1$ with an infinitely large size must be of the form
$f(A_1^\infty,g(A_2^\infty))$ where $A_1^\infty$ or $A_2^\infty$ or both
are of the form $f(A_1^\infty,g(A_2^\infty))$. 
We now consider the following two cases.

\begin{enumerate}
\item
No $L_{i_j}^1$ in $D$ contains elements with 
an infinitely large size. Let $N$ be the largest size of an element
$f(\_,g(\_))$ in the $L_{i_j}^1$s and let $E_2=\{f(A_1^k,g(A_2^k))|k\geq 1$ such that
$|f(A_1^k,g(A_2^k))|\leq N\}$. Obviously $E_2$ is finite. So the elements of any 
$L_{i_j}^1$ in $D$ come from $E\cup E_2$. Since $E\cup E_2$ is finite, the
number of the combinations of elements of $E\cup E_2$ is finite. 
Since the $L_{i_j}^1$s in $D$ grow towards an infinitely large size, 
some combinations must repeat in $D$ infinite times. 
This suggests that no matter what shuffling approach is used, there must be 
an infinite sequence 
$L_{g_1}^1,L_{g_2}^1...$ among the $L_{i_j}^1$s
such that for any $k \geq 1$, let $L_{g_k}^1=p([S_1,...,S_n])$,
then after variable renaming $L_{g_{k+1}}^1$ will become like 
$p([\vec{S_1},...,\vec{S_n}])$ such that each $S_l$
is in $\vec{S_l}$. These
$S_l$s in $L_{g_k}^1$ are growing terms w.r.t. $L_{g_{k+1}}^1$, thus
$L_{g_{k+1}}^1\sqsupseteq_{EV}L_{g_k}^1$ with $|L_{g_{k+1}}^1|>|L_{g_k}^1|$.

\item
As $j\rightarrow \infty$, 
some elements in $L_{i_j}^1$ grow towards an infinitely large size.
Let $T^{j+1}$ be an element in the list of $L_{i_{j+1}}^1$ that (as
$j\rightarrow \infty$) grows towards an element with an infinitely large
size. Then there must be an element $T^j$ in the list of $L_{i_j}^1$
that grows towards an element with an infinitely large
size via $T^{j+1}$ (otherwise, $T^{j+1}$ would not grow towards an
infinitely large element since we apply the same set $S$ of clauses
from $L_{i_{j+1}}^1$ to $L_{i_{j+2}}^1$ as from $L_{i_j}^1$ to $L_{i_{j+1}}^1$).
This means that $T^{j+1}$ is the same as (or a variant of) $T^j$ or 
$T^{j+1}=f(A_1^j,g(A_2^j))$ such that $T^j$ (or its variant) is in
$A_1^j$ or $A_2^j$. Obviously $T^{j+1}$ is an expanded variant of $T^j$.
Generalizing such argument, for each infinitely large element $T_l^\infty$
in the list of $L_{i_\infty}^1$ we have an infinite sequence
\begin{tabbing}
$\qquad$ $T^1,T^2,...,T^j,T^{j+1}, ..., T_l^\infty$
\end{tabbing}
with each $T^j$ in the list of 
$L_{i_j}^1$ such that $T^1$ grows towards $T_l^\infty$
via $T^2$ that grows towards $T_l^\infty$ via $T^3$, and so on. Obviously,
for any $k>j\geq 1$ $T^k$ is an expanded variant of $T^j$. So in this case
each $T^j$ is called a {\em growing element} w.r.t. $T_l^\infty$. Note that
if there are more than one element in some $L_{i_j}^1$ that grow towards $T_l^\infty$
via $T^{j+1}$, such as in the case $T^{j+1}=f(T_1^j,g(T_2^j))$ with $T_1^j,T_2^j$ in
$L_{i_j}^1$, only one $T^j$ of them is selected as the growing element w.r.t.
$T_l^\infty$ based on the following criterion: $T^j$ must be an expanded variant
of $T^{j-1}$. If still more than one element meet such criterion, select
an arbitrary one.

\hspace{.15in} We now partition the list $[T_1^j,...,T_{m_j}^j]$ of
each $L_{i_j}^1$ into two parts: the sublist $GE_{i_j}$ of growing
elements and the sublist $NGE_{i_j}$ of non-growing elements.
That is, $T\in [T_1^j,...,T_{m_j}^j]$ is in $GE_{i_j}$ if it is a 
growing element w.r.t. some $T_l^\infty$. Clearly, for each $k>j\geq 1$
$|GE_{i_k}|\geq |GE_{i_j}|$. Since for any $V_1,...,V_m$ in $GE_{i_j}$
there are $m$ elements $V_1',...,V_m'$ in $GE_{i_k}$ such that each
$V_l'$ is an expanded variant of $V_l$, there must be an infinite
sequence $L_{f_1}^1,L_{f_2}^1, ...$ among the $L_{i_j}^1$s such that for 
any $k>j\geq 1$ $GE_{f_k}$ is an expanded variant of $GE_{f_j}$.
That is, let $GE_{f_j}=[V_1,...,V_m]$, then after variable renaming
$GE_{f_k}$ becomes $[\vec{V_1},...,\vec{V_m}]$ such that each $\vec{V_l}$
contains an element that is an expanded variant of $V_l$.

\hspace{.15in} Now consider the elements of $L_{f_j}^1$s. Since all infinitely
large elements have been covered by the growing elements, the size
of any non-growing element is bounded by some constant, say $N$. 
Let $E_2=\{f(A_1^k,g(A_2^k))|k\geq 1$ such that
$|f(A_1^k,g(A_2^k))|\leq N\}$. So all non-growing elements of any 
$L_{f_j}^1$ come from $E\cup E_2$. Since $E\cup E_2$ is finite and
the sequence $L_{f_1}^1,L_{f_2}^1, ...$ is infinite,
some combinations of elements of $E\cup E_2$ must occur in infinitely many $L_{f_j}^1$s. 
This implies that no matter what shuffling approach is used, there must be 
an infinite sequence $L_{g_1}^1,L_{g_2}^1...$ among the $L_{f_j}^1$s
such that for any $k \geq 1$, let $L_{g_k}^1=p([S_1,...,S_n])$,
then after variable renaming $L_{g_{k+1}}^1$ will become like 
$p([\vec{S_1},...,\vec{S_n}])$ such that each $\vec{S_l}$ contains
an element that is an expanded variant of $S_l$. That is,
$L_{g_{k+1}}^1\sqsupseteq_{EV}L_{g_k}^1$ with $|L_{g_{k+1}}^1|>|L_{g_k}^1|$.     

\end{enumerate}

\end{enumerate}

\noindent {\bf Proof of Theorem \ref{th3}.}
By the proof of Theorem \ref{th2}, $D$ contains an infinite
number of selected subgoals $L_1^1,L_2^1,...$ such that 
$L_j^1 \prec_{ANC} L_{j+1}^1$ $(j\geq 1)$.
Since any logic program has only a finite number of clauses, there must
be a set of clauses in the program that are invoked an infinite number of 
times in $D$. Let $S=\{C_1, ..., C_n\}$ be the set of all different
clauses that are used an infinite number of times in $D$. Then $D$ can
be depicted as
\begin{tabbing}
$\qquad$ $N_0:G_0\Rightarrow ... N_{i_1}:G_{i_1} 
\stackrel{C_{1_1}}{\longmapsto} ... \stackrel{C_{1_{n_1}}}{\longmapsto}
N_{i_2}:G_{i_2} 
\stackrel{C_{2_1}}{\longmapsto} ... \stackrel{C_{2_{n_2}}}{\longmapsto}
N_{i_3}:G_{i_3} 
\stackrel{C_{3_1}}{\longmapsto} ...$
\end{tabbing}
where for any $j \geq 1$, $L_{i_j}^1\prec_{ANC}L_{i_{j+1}}^1$
and  $\{C_{j_1}, ..., C_{j_{n_j}}\}=S$.
Since any logic program has only a finite 
number of predicate, function and constant
symbols and $D$ contains subgoals with infinitely large size, 
there must be an infinite sequence
$L_{f_1}^1,L_{f_2}^1...$ among the $L_{i_j}^1$s such that
for any $l \geq 1$, 
$L_{f_l}^1$ is a variant of $L_{f_{l+1}}^1$ 
except for a few terms in $L_{f_{l+1}}^1$ that
increase in size. Hence by Lemma \ref{lemm31}, there is 
an infinite sequence
$L_{g_1}^1,L_{g_2}^1...$ among the $L_{f_l}^1$s
such that for any $k\geq 1$ $L_{g_{k+1}}^1\sqsupseteq_{EV} L_{g_k}^1$ 
with $|L_{g_{k+1}}^1|>|L_{g_k}^1|$.   

\begin{theorem}
\label{th4}
$D$ is an infinite generalized SLDNF-derivation 
if and only if it is of the form
\begin{tabbing}
$\qquad$ $N_0:G_0\Rightarrow ... N_{g_1}:G_{g_1} 
\stackrel{C_{1_1}}{\longmapsto} ... \stackrel{C_{1_{n_1}}}{\longmapsto}
N_{g_2}:G_{g_2} 
\stackrel{C_{2_1}}{\longmapsto} ... \stackrel{C_{2_{n_2}}}{\longmapsto}
N_{g_3}:G_{g_3} 
\stackrel{C_{3_1}}{\longmapsto} ...$
\end{tabbing}
such that 
\begin{enumerate}
\item
For any $j \geq 1$, $L_{g_j}^1\prec_{ANC}L_{g_{j+1}}^1$ and
$L_{g_{j+1}}^1 \sqsupseteq_{EV} L_{g_j}^1$.

\item
For any $j \geq 1$ $|L_{g_j}^1|=|L_{g_{j+1}}^1|$, or
for any $j \geq 1$ $|L_{g_j}^1|<|L_{g_{j+1}}^1|$.

\item
For any $j \geq 1$, the set of clauses used to derive
$L_{g_{j+1}}^1$ from $L_{g_j}^1$ is the same as that of deriving
$L_{g_{j+2}}^1$ from $L_{g_{j+1}}^1$, i.e.
$\{C_{j_1},...,C_{j_{n_j}}\}=\{C_{(j+1)_1},...,C_{(j+1)_{n_{j+1}}}\}$.
\end{enumerate}
\end{theorem}

\noindent {\bf Proof.}
$(\Longleftarrow)$ Straightforward.

$(\Longrightarrow)$
By Theorems \ref{th2} and \ref{th3}, $D$ is of the form
\begin{tabbing}
$\qquad$ $N_0:G_0\Rightarrow ... \Rightarrow N_{i_1}:G_{i_1} \Rightarrow ... \Rightarrow
N_{i_2}:G_{i_2}\Rightarrow ... $
\end{tabbing}
where for any $j \geq 1$, $L_{i_j}^1\prec_{ANC}L_{i_{j+1}}^1$ and
$L_{i_{j+1}}^1\sqsupseteq_{EV}L_{i_j}^1$. In particular, when all subgoals in $D$
are bounded in size, by Theorem \ref{th2} for any $j \geq 1$
$|L_{i_j}^1|=|L_{i_{j+1}}^1|$. Otherwise, by Theorem \ref{th3} for any $j \geq 1$
$|L_{i_j}^1|<|L_{i_{j+1}}^1|$.

Since any logic program has only a finite number of clauses, there must
be a set $S=\{C_1, ..., C_n\}$ of clauses in the program that are 
invoked an infinite number of times in $D$. This means that there exists an 
infinite sequence of subgoals $L_{g_1}^1,L_{g_2}^1,...$ among the $L_{i_j}^1$s
such that for any $j\geq 1$ $L_{g_{j+1}}^1$ is derived from $L_{g_j}^1$
by applying the set $S$ of clauses. That is, $D$ is of the form
\begin{tabbing}
$\qquad$ $N_0:G_0\Rightarrow ... N_{g_1}:G_{g_1} 
\stackrel{C_{1_1}}{\longmapsto} ... \stackrel{C_{1_{n_1}}}{\longmapsto}
N_{g_2}:G_{g_2} 
\stackrel{C_{2_1}}{\longmapsto} ... \stackrel{C_{2_{n_2}}}{\longmapsto}
N_{g_3}:G_{g_3} 
\stackrel{C_{3_1}}{\longmapsto} ...$
\end{tabbing}
such that the three conditions of this theorem hold. \\

Theorem \ref{th4} is the principal result of this paper. It captures
two crucial characteristics of an infinite generalized SLDNF-derivation:
repetition of selected subgoals and clauses, and recursive
increase in term size. Repetition leads to 
variants, whereas recursive increase introduces growing terms.
It is the characterization of these key 
(dynamic) features that allows us to design a mechanism
for automatically testing termination of general logic programs.

\section{Testing Termination of General Logic Programs}
\label{verify}
\subsection{Definition of Termination}
In \cite{DD93}, a generic definition of termination of
logic programs is presented.

\begin{definition}
\label{term-def1}
{\em
Let $P$ be a general logic program, $S_Q$ a set of queries and $S_R$
a set of selection rules. $P$ is terminating w.r.t. $S_Q$ and $S_R$
if for each query $Q_i$ in $S_Q$ and for each selection rule $R_j$ in $S_R$, 
all SLDNF-trees for $P\cup \{\leftarrow Q_i\}$ via $R_j$ are finite. 
}
\end{definition}

Observe that the above definition considers finite SLDNF-trees 
for termination. That is, if $P$ is terminating w.r.t. $Q_i$ then all (complete)
SLDNF-trees for $P\cup \{\leftarrow Q_i\}$ must be finite. This does not
seem to apply to Prolog where there exist cases in which, although $P$ is terminating
w.r.t. $Q_i$ and $R_j$, some (complete) SLDNF-trees for $P\cup \{\leftarrow Q_i\}$ are
infinite. Example \ref{eg2} gives such an illustration, where Prolog
terminates with a positive answer.

In view of the above observation, we present the following definition based on 
a generalized SLDNF-tree.

\begin{definition}
\label{term-def2}
{\em
Let $P$ be a general logic program, $S_Q$ a finite set of queries and $R$
a fixed depth-first, left-most control strategy. 
$P$ is terminating w.r.t. $S_Q$ and $R$
if for each query $Q_i$ in $S_Q$, the generalized SLDNF-tree
for $P\cup \{\leftarrow Q_i\}$ via $R$ is finite. 
}
\end{definition}

The above definition implies that $P$ is terminating w.r.t. $S_Q$ and $R$ 
if and only if there is no infinite generalized SLDNF-derivation
in any generalized SLDNF-tree $GT_{\leftarrow Q_i}$. So
the following result is immediate from Theorem \ref{th4}.

\begin{theorem}
\label{th-iff}
$P$ is terminating w.r.t. $S_Q$ and $R$ if and only if for each query $Q_i$ in $S_Q$
there is no infinite generalized SLDNF-derivation 
in $GT_{\leftarrow Q_i}$ of the form
\begin{tabbing}
$\qquad$ $N_0:G_0\Rightarrow ... N_{g_1}:G_{g_1} 
\stackrel{C_{1_1}}{\longmapsto} ... \stackrel{C_{1_{n_1}}}{\longmapsto}
N_{g_2}:G_{g_2} 
\stackrel{C_{2_1}}{\longmapsto} ... \stackrel{C_{2_{n_2}}}{\longmapsto}
N_{g_3}:G_{g_3} 
\stackrel{C_{3_1}}{\longmapsto} ...$
\end{tabbing}
that meets the three conditions of Theorem \ref{th4}.
\end{theorem}

\subsection{An Algorithm for Automatically Testing Termination}
Theorem \ref{th-iff} provides a necessary and sufficient condition
for termination of a general logic program. Obviously, such a condition cannot
be directly used for automatic verification because it requires generating
an infinite generalized SLDNF-derivation to see if the three conditions of
Theorem \ref{th4} are satisfied.

As we mentioned before, the three conditions of Theorem \ref{th4} capture two most important
structural features of an infinite generalized SLDNF-derivation. Therefore,
we may well use these conditions to predict possible infinite generalized SLDNF-derivations
based on some finite generalized SLDNF-derivations. Although the predictions may not
always be guaranteed to be correct (since the termination problem
is undecidable in general), it can be correct in a vast majority of cases.
That is, if the three conditions of Theorem \ref{th4} are satisfied by some
finite generalized SLDNF-derivation, the underlying general
logic program is most likely non-terminating. This leads to the
following definition.

\begin{definition}
\label{most-non-term}
{\em
Let $P$ be a general logic program, $S_Q$ a finite set of queries and $R$
a depth-first, left-most control strategy. Let $d>1$ be a depth bound.
$P$ is said to be {\em most-likely} non-terminating w.r.t. $S_Q$ and $R$
if for some query $Q_i$ in $S_Q$, there is a generalized SLDNF-derivation
of the form
\begin{tabbing}
$\qquad$ $N_0:\leftarrow Q_i\Rightarrow ... $ \= $N_{g_1}:G_{g_1} 
\stackrel{C_{1_1}}{\longmapsto} ... \stackrel{C_{1_{n_1}}}{\longmapsto}$\\
\> $N_{g_2}:G_{g_2} 
\stackrel{C_{2_1}}{\longmapsto} ... \stackrel{C_{2_{n_2}}}{\longmapsto}$\\
\> $\qquad\quad$ \vdots\\
\> $N_{g_d}:G_{g_d} 
\stackrel{C_{d_1}}{\longmapsto} ... \stackrel{C_{d_{n_d}}}{\longmapsto}
N_{g_{d+1}}:G_{g_{d+1}}$
\end{tabbing}
such that 
\begin{enumerate}
\item
For any $j \leq d$, $L_{g_j}^1\prec_{ANC}L_{g_{j+1}}^1$ and
$L_{g_{j+1}}^1 \sqsupseteq_{EV} L_{g_j}^1$.

\item
For any $j \leq d$ $|L_{g_j}^1|=|L_{g_{j+1}}^1|$, or
for any $j \leq d$ $|L_{g_j}^1|<|L_{g_{j+1}}^1|$.

\item
For any $j \leq d$, the set of clauses used to derive
$L_{g_{j+1}}^1$ from $L_{g_j}^1$ is the same as that of deriving
$L_{g_{j+2}}^1$ from $L_{g_{j+1}}^1$, i.e.
$\{C_{j_1},...,C_{j_{n_j}}\}=\{C_{(j+1)_1},...,C_{(j+1)_{n_{j+1}}}\}$.
\end{enumerate}
 
}
\end{definition}

\begin{theorem}
\label{th-likely}
Let $P$, $S_Q$ and $R$ be as defined in Definition \ref{most-non-term}.
\begin{enumerate}
\item
\label{it-11}
If $P$ is not terminating w.r.t. $S_Q$ and $R$ then it is most-likely
non-terminating w.r.t. $S_Q$ and $R$. 

\item
\label{it-12}
If $P$ is not most-likely non-terminating w.r.t. $S_Q$ and $R$ then
it is terminating w.r.t. $S_Q$ and $R$.
\end{enumerate}
\end{theorem}

\noindent {\bf Proof:} \ref{it-11}.
If $P$ is not terminating w.r.t. $S_Q$ and $R$, by Definition \ref{term-def2} 
for some query $Q_i\in S_Q$ there exists an infinite generalized
SLDNF-derivation in $GT_{\leftarrow Q_i}$. 
The result is then immediate from Theorem \ref{th4}.

\ref{it-12}. 
If $P$ is not most-likely non-terminating w.r.t. $S_Q$ and $R$ and, on the contrary,
it is not terminating w.r.t. $S_Q$ and $R$, then by the first part of this theorem
we reach a contradiction.\\

It is easily seen that the converse of the above theorem does not hold.
The following algorithm is to determine most-likely non-termination.

\begin{algorithm}
\label{alg1}
{\em
Testing termination of a general logic program.
\begin{itemize}
\item
Input: A general logic program $P$, a finite set of queries
$S_Q=\{Q_1,...,Q_m\}$, and a depth-first, left-most control strategy $R$.
\item
Output: {\em Yes} or a generalized SLDNF-derivation $D$.
\item
Method: Apply the following procedure.
\begin{tabbing}
{\em procedure} $Test(P,S_Q,R)$\\
$\quad$ \= {\em begin}\\
1 \> $\quad$ \= For each query $Q_i\in S_Q$, construct the full generalized SLDNF-tree\\
\>\> $GT_{\leftarrow Q_i}$ for $P\cup \{\leftarrow Q_i\}$ via $R$ unless a generalized 
SLDNF-derivation \\
\>\> $D$ is encountered that meets the three conditions of 
Definition \ref{most-non-term},\\
\>\> in which case return $D$ and stop the procedure;\\
2 \>\> Return {\em Yes}\\
\> {\em end}
\end{tabbing}

\end{itemize}
}
\end{algorithm}

\begin{theorem}
\label{th5}
Algorithm \ref{alg1} terminates. It returns {\it Yes} if and only if
$P$ is not most-likely non-terminating w.r.t. $S_Q$ and $R$.
\end{theorem}

\noindent {\bf Proof:}
If for each query $Q_i\in S_Q$ the generalized SLDNF-tree 
$GT_{\leftarrow Q_i}$ for $P\cup\{\leftarrow Q_i\}$
is finite, line 1 of Algorithm \ref{alg1} will be completed
in finite time, so that Algorithm \ref{alg1} will terminate in finite time.
Otherwise, let all generalized SLDNF-trees $GT_{\leftarrow Q_i}$
with $i<m$ be finite and $GT_{\leftarrow Q_{i+1}}$ be infinite. Let
$D$ be the first infinite derivation in $GT_{\leftarrow Q_{i+1}}$.
By Theorem \ref{th4}, $D$ must be of the form
\begin{tabbing}
$\qquad$ $N_0:G_0\Rightarrow ... N_{g_1}:G_{g_1} 
\stackrel{C_{1_1}}{\longmapsto} ... \stackrel{C_{1_{n_1}}}{\longmapsto}
N_{g_2}:G_{g_2} 
\stackrel{C_{2_1}}{\longmapsto} ... \stackrel{C_{2_{n_2}}}{\longmapsto}
N_{g_3}:G_{g_3} 
\stackrel{C_{3_1}}{\longmapsto} ...$
\end{tabbing}
such that the three conditions of Theorem \ref{th4} hold. Obviously,
such an infinite derivation will be detected at the node 
$N_{g_{d+1}}:G_{g_{d+1}}$, thus Algorithm \ref{alg1} will stop here.

When Algorithm \ref{alg1} ends with an answer {\em Yes}, all generalized
SLDNF-trees for all queries in $S_Q$ must have been generated without
encountering any derivation $D$ that meets the three conditions 
of Definition \ref{most-non-term}. This shows that $P$
is not most-likely non-terminating w.r.t. $S_Q$ and $R$.
Conversely, if $P$ is not most-likely non-terminating w.r.t. $S_Q$ and $R$,
Algorithm \ref{alg1} will not stop at line 1. It will proceed to line 2
with an answer {\em Yes} returned.

\begin{theorem}
\label{th6}
The following hold:
\begin{enumerate}
\item
\label{it-21}
If Algorithm \ref{alg1} returns {\it Yes} then $P$ is terminating
w.r.t. $S_Q$ and $R$.

\item
\label{it-22}
If $P$ is not terminating w.r.t. $S_Q$ and $R$ 
then Algorithm \ref{alg1} will return
a generalized SLDNF-derivation $D$ that meets the conditions of 
Definition \ref{most-non-term}.
\end{enumerate}
\end{theorem}

\noindent {\bf Proof:}
\ref{it-21}. By Theorem \ref{th5} Algorithm \ref{alg1} 
returning {\em Yes} implies $P$ is not
most-likely non-terminating w.r.t. $S_Q$ and $R$. The result then follows 
from Theorem \ref{th-likely}.

\ref{it-22}. By Theorem \ref{th-likely}, when $P$ is not
terminating w.r.t. $S_Q$ and $R$, it is
most-likely non-terminating w.r.t. $S_Q$ and $R$. So there exist generalized
SLDNF-derivations in some generalized SLDNF-trees $GT_{\leftarrow Q_i}$
that meet the three conditions of Definition \ref{most-non-term}.
Obviously, the fist such derivation $D$ will be captured at line 1 of 
Algorithm \ref{alg1}, which leads to an output $D$. 

\subsection{Examples}
We use the following very representative examples to illustrate
the effectiveness of our method. In the sequel, we choose the smallest 
depth bound $d=2$.

\begin{example}
\label{eg3}
{\em
Applying Algorithm \ref{alg1} to the logic program $P_1$ of
Example \ref{eg1} with a query
$Q_1=p(a)$ will return a generalized SLDNF-derivation $D$,
which is the path from $N_0$ to $N_4$ in Figure \ref{fig1}.
$D$ is informative enough to suggest that $P_1$ is not terminating
w.r.t. $Q_1$.
}
\end{example}

\begin{example}
\label{eg4}
{\em
Applying Algorithm \ref{alg1} to the logic program $P_2$ of
Example \ref{eg2} with a query
$Q_1=p$ will return an answer {\it Yes}. That is, $P_2$
is terminating w.r.t. $Q_1$. However, for the query $Q_2=q$
applying Algorithm \ref{alg1} will return the following generalized 
SLDNF-derivation 
\begin{tabbing}
$\qquad$ $N_0:\leftarrow q\stackrel{C_{q_2}}{\longrightarrow}
N_1:\leftarrow q\stackrel{C_{q_2}}{\longrightarrow}N_2:\leftarrow q$
\end{tabbing}
which strongly suggests that $P_2$ is not terminating w.r.t. $Q_2$.
}
\end{example}

\begin{example}
\label{eg5}
{\em
Consider the following widely used program:\footnote{It represents
a large class of well-known logic programs such as $member$, $subset$,
$merge$, $quick$-$sort$, $reverse$, $permutation$, and so on.}
 
\begin{tabbing}
$\qquad$ \= $P_3:$ $\quad$ \= $append([],X,X)$. \`$C_{a_1}$\\
\>\> $append([X|Y],U,[X|Z])\leftarrow append(Y,U,Z)$.       \`$C_{a_2}$ 
\end{tabbing}
Assume the following types of queries (borrowed from \cite{DD93}): 
\begin{tabbing}
$\qquad$ \= $Q_1= append([1,2],[3],L)$,\\
\> $Q_2=append([1,2],[3],[4])$,\\
\> $Q_3=append(L_1,L_2,[1,2])$,\\
\> $Q_4=append(L_1,[1,2],L_3)$,\\
\> $Q_5=append(L_1,L_2,L_3)$,\\
\> $Q_6=append([X|Y],[],Y)$,\\
\> $Q_7=append([X|Y],Y,[Z|Y])$.
\end{tabbing}

The generalized SLDNF-trees $GT_{\leftarrow Q_1}$,
$GT_{\leftarrow Q_2}$, and $GT_{\leftarrow Q_3}$ are all finite
and contain no expanded variants in any derivations. So
Algorithm \ref{alg1} will return {\em Yes} when executing
$Test(P_3,\{Q_1,Q_2,Q_3\},R)$. That is, $P_3$
is terminating w.r.t. the first three types of queries.

When evaluating $Q_4$, Algorithm \ref{alg1} will return
a generalized SLDNF-derivation as shown in Figure \ref{fig3} (a). Note that
all selected subgoals in the derivation are variants. Similar derivations
will be returned when applying Algorithm \ref{alg1} to 
$Q_5$ and $Q_6$. Applying Algorithm \ref{alg1} to $Q_7$
will yield a generalized SLDNF-derivation as shown in Figure \ref{fig3} (b).
Note that the selected subgoal at node $N_3$ is an expanded variant
of the subgoal at $N_2$ that is an expanded variant of the subgoal
at $N_1$. That is, 
$append(Y_2,[X_1|[X_2|Y_2]],Y_2)\sqsupseteq_{EV}
append(Y_1,[X_1|Y_1],Y_1)\sqsupseteq_{EV}
append(Y,Y,Y)$.

It is clear that the derivations of Figures \ref{fig3} (a) and (b)
can be infinitely extended by repeatedly applying the clause
$C_{a_2}$, thus $P_3$ is non-terminating w.r.t. the queries $Q_4-Q_7$.

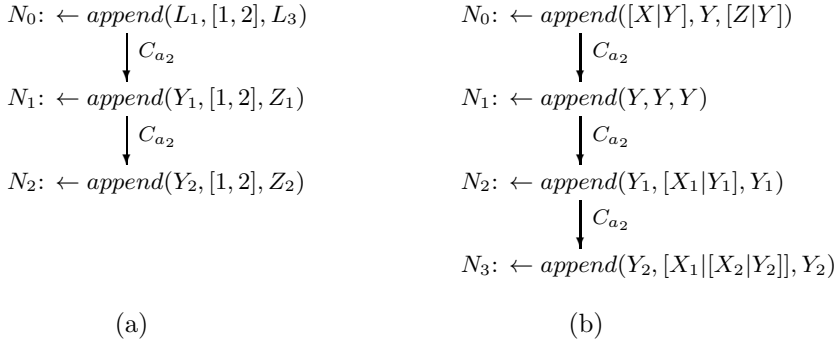
\begin{figure}[h]
$\qquad\quad$
\setlength{\unitlength}{3947sp}%
\begingroup\makeatletter\ifx\SetFigFont\undefined%
\gdef\SetFigFont#1#2#3#4#5{%
  \reset@font\fontsize{#1}{#2pt}%
  \fontfamily{#3}\fontseries{#4}\fontshape{#5}%
  \selectfont}%
\fi\endgroup%
\begin{picture}(3675,2046)(451,-1336)
\put(1126,-1336){\makebox(0,0)[lb]{\smash{\SetFigFont{10}{12.0}{\rmdefault}{\mddefault}{\updefault}(a)}}}
\put(3976,-1336){\makebox(0,0)[lb]{\smash{\SetFigFont{10}{12.0}{\rmdefault}{\mddefault}{\updefault}(b)}}}
\thinlines
\put(1201,539){\vector( 0,-1){300}}
\put(1201, 14){\vector( 0,-1){300}}
\put(4051,539){\vector( 0,-1){300}}
\put(4051, 14){\vector( 0,-1){300}}
\put(4051,-511){\vector( 0,-1){300}}
\put(1276,389){\makebox(0,0)[lb]{\smash{\SetFigFont{8}{9.6}{\rmdefault}{\mddefault}{\updefault}$C_{a_2}$}}}
\put(1276,-136){\makebox(0,0)[lb]{\smash{\SetFigFont{8}{9.6}{\rmdefault}{\mddefault}{\updefault}$C_{a_2}$}}}
\put(451,614){\makebox(0,0)[lb]{\smash{\SetFigFont{9}{10.8}{\rmdefault}{\mddefault}{\updefault}$N_0$:  $\leftarrow append(L_1,[1,2],L_3)$}}}
\put(451, 89){\makebox(0,0)[lb]{\smash{\SetFigFont{9}{10.8}{\rmdefault}{\mddefault}{\updefault}$N_1$:  $\leftarrow append(Y_1,[1,2],Z_1)$ }}}
\put(451,-436){\makebox(0,0)[lb]{\smash{\SetFigFont{9}{10.8}{\rmdefault}{\mddefault}{\updefault}$N_2$:  $\leftarrow append(Y_2,[1,2],Z_2)$ }}}
\put(4126,389){\makebox(0,0)[lb]{\smash{\SetFigFont{8}{9.6}{\rmdefault}{\mddefault}{\updefault}$C_{a_2}$}}}
\put(4126,-136){\makebox(0,0)[lb]{\smash{\SetFigFont{8}{9.6}{\rmdefault}{\mddefault}{\updefault}$C_{a_2}$}}}
\put(3301,614){\makebox(0,0)[lb]{\smash{\SetFigFont{9}{10.8}{\rmdefault}{\mddefault}{\updefault}$N_0$:  $\leftarrow append([X|Y],Y,[Z|Y])$}}}
\put(3301, 89){\makebox(0,0)[lb]{\smash{\SetFigFont{9}{10.8}{\rmdefault}{\mddefault}{\updefault}$N_1$:  $\leftarrow append(Y,Y,Y)$ }}}
\put(4126,-661){\makebox(0,0)[lb]{\smash{\SetFigFont{8}{9.6}{\rmdefault}{\mddefault}{\updefault}$C_{a_2}$}}}
\put(3301,-436){\makebox(0,0)[lb]{\smash{\SetFigFont{9}{10.8}{\rmdefault}{\mddefault}{\updefault}$N_2$:  $\leftarrow append(Y_1,[X_1|Y_1],Y_1)$ }}}
\put(3301,-961){\makebox(0,0)[lb]{\smash{\SetFigFont{9}{10.8}{\rmdefault}{\mddefault}{\updefault}$N_3$:  $\leftarrow append(Y_2,[X_1|[X_2|Y_2]],Y_2)$ }}}
\end{picture}

\caption{Two generalized SLDNF-derivations that satisfy the three conditions of 
Definition \ref{most-non-term}}\label{fig3}
\end{figure}
}
\end{example}

\begin{example}
\label{eg6-0}
{\em
The following program illustrates how a list of terms grows
recursively through a DELETE-ADD-SHUFFLE process.
\begin{tabbing}
$\qquad$ \= $P_4:$ $\quad$ \= $p([X_1,X_2|Y])\leftarrow 
   q([X_1,X_2|Y],Z),reverse(Z,[],Z_1),p(Z_1)$. \`$C_{p_1}$\\
\>\> $q([X_1,X_2|Y],[X_3,f(X_1,X_2),X_2|Y]).$       \`$C_{q_1}$ \\
\>\> $reverse(Z,[],Z_1)\leftarrow$ $Z_1$ is the reversed list of $Z$.  \`$C_{rev}$ 
\end{tabbing}
Given a subgoal $p([X_1,X_2|Y])$,
applying $C_{p_1},C_{q_1},C_{rev}$ successively will
\begin{tabbing}
$\qquad$ \= DELETE $\quad$ \= $X_1$, thus yielding a list $[X_2|Y]$,\\
\> ADD \> $X_3$ and $f(X_1,X_2)$, thus yielding a list $[X_3,f(X_1,X_2),X_2|Y]$, and\\
\> SHUFFLE \> the list $[X_3,f(X_1,X_2),X_2|Y]$ by reversing its components.
\end{tabbing}
Note that the addition of $X_3$ is independent of the original list $[X_1,X_2|Y]$,
but $f(X_1,X_2)$ is generated based on the list. This means that given a query of the 
form $p([T_1,...,T_m])$, a new variable $X$ and
a function $f(A_1^j,A_2^j)$ will be added each cycle $\{C_{p_1},C_{q_1},C_{rev}\}$
is applied, where the domain of $A_1^j$ and $A_2^j$ is the closure of the function
$f(\_,\_)$ over $\{X,T_1,...,T_m\}$ (up to variable renaming). As an illustration,
consider an arbitrary query $Q_1=p([a,b])$.
Applying Algorithm \ref{alg1} to $Q_1$
will return a generalized SLDNF-derivation as shown in Figure \ref{fig4-0},
where for the selected subgoals $L_{12}^1,L_6^1,L_0^1$ 
at nodes $N_{12}$, $N_6$ and $N_0$, we have  
$L_0^1\prec_{ANC}L_6^1\prec_{ANC}L_{12}^1$,
$L_{12}^1 \sqsupseteq_{EV} L_6^1\sqsupseteq_{EV} L_0^1$, and
$|L_{12}^1|> |L_6^1|>|L_0^1|$. 
We see the following terms added due to the repeated applications 
of $\{C_{p_1},C_{q_1},C_{rev}\}$:
\begin{tabbing}
$\qquad$ \=  From \= $N_0$ to $N_3$ $\qquad\quad$ \= $X_1, f(a,b),$\\
\> \>   $N_3$ to $N_6$  \> $X_2, f(b,f(a,b)),$\\
\> \>   $N_6$ to $N_9$  \> $X_3, f(X_1,f(a,b)),$\\
\> \>   $N_9$ to $N_{12}$  \> $X_4, f(X_2,f(b,f(a,b))).$
\end{tabbing}

Apparently, the generalized SLDNF-derivation of Figure \ref{fig4-0} 
can be infinitely extended. Thus
$P_4$ is non-terminating w.r.t. $Q_1$ (and all queries of the 
form $p([T_1,...,T_m])$). 
 
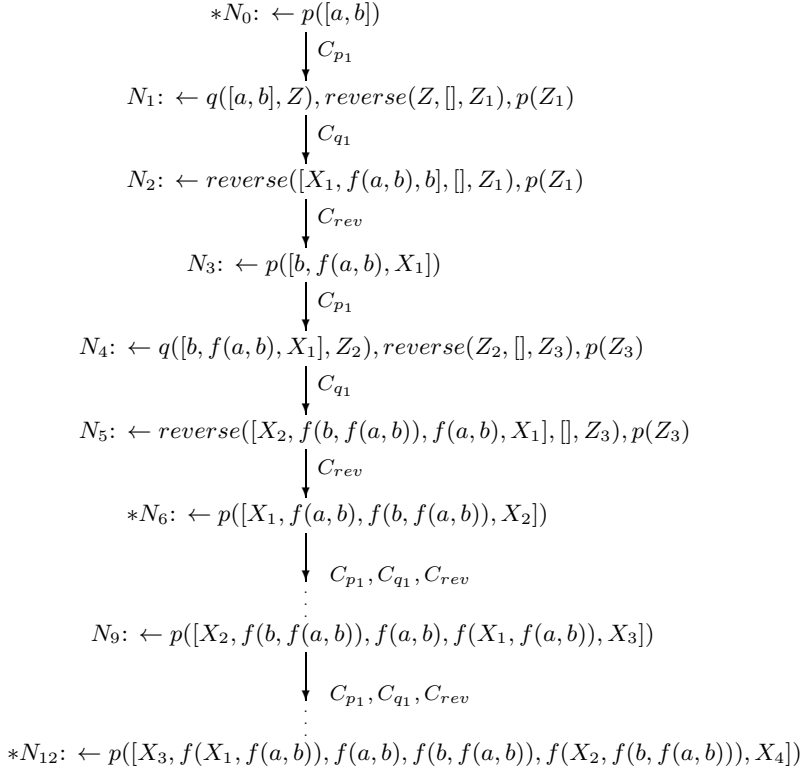
\begin{figure}[htb]
$\qquad\quad$
\setlength{\unitlength}{3947sp}%
\begingroup\makeatletter\ifx\SetFigFont\undefined%
\gdef\SetFigFont#1#2#3#4#5{%
  \reset@font\fontsize{#1}{#2pt}%
  \fontfamily{#3}\fontseries{#4}\fontshape{#5}%
  \selectfont}%
\fi\endgroup%
\begin{picture}(2025,4782)(2701,-4072)
\thinlines
\multiput(4576,-2986)(0.00000,-75.00000){3}{\makebox(1.6667,11.6667){\SetFigFont{5}{6}{\rmdefault}{\mddefault}{\updefault}.}}
\multiput(4576,-3736)(0.00000,-75.00000){3}{\makebox(1.6667,11.6667){\SetFigFont{5}{6}{\rmdefault}{\mddefault}{\updefault}.}}
\put(4576,539){\vector( 0,-1){300}}
\put(4576,-1036){\vector( 0,-1){300}}
\put(4576,-1561){\vector( 0,-1){300}}
\put(4576, 14){\vector( 0,-1){300}}
\put(4576,-511){\vector( 0,-1){300}}
\put(4576,-2086){\vector( 0,-1){300}}
\put(4576,-2611){\vector( 0,-1){300}}
\put(4576,-3361){\vector( 0,-1){300}}
\put(3976,614){\makebox(0,0)[lb]{\smash{\SetFigFont{9}{10.8}{\rmdefault}{\mddefault}{\updefault}$*N_0$:  $\leftarrow p([a,b])$}}}
\put(4651,389){\makebox(0,0)[lb]{\smash{\SetFigFont{8}{9.6}{\rmdefault}{\mddefault}{\updefault}$C_{p_1}$}}}
\put(3451, 89){\makebox(0,0)[lb]{\smash{\SetFigFont{9}{10.8}{\rmdefault}{\mddefault}{\updefault}$N_1$:  $\leftarrow q([a,b],Z),reverse(Z,[],Z_1),p(Z_1)$}}}
\put(4651,-661){\makebox(0,0)[lb]{\smash{\SetFigFont{8}{9.6}{\rmdefault}{\mddefault}{\updefault}$C_{rev}$}}}
\put(3451,-436){\makebox(0,0)[lb]{\smash{\SetFigFont{9}{10.8}{\rmdefault}{\mddefault}{\updefault}$N_2$:  $\leftarrow reverse([X_1,f(a,b),b],[],Z_1),p(Z_1)$}}}
\put(3451,-2536){\makebox(0,0)[lb]{\smash{\SetFigFont{9}{10.8}{\rmdefault}{\mddefault}{\updefault}$*N_6$:  $\leftarrow p([X_1,f(a,b),f(b,f(a,b)),X_2])$}}}
\put(4651,-136){\makebox(0,0)[lb]{\smash{\SetFigFont{8}{9.6}{\rmdefault}{\mddefault}{\updefault}$C_{q_1}$}}}
\put(4651,-1186){\makebox(0,0)[lb]{\smash{\SetFigFont{8}{9.6}{\rmdefault}{\mddefault}{\updefault}$C_{p_1}$}}}
\put(4651,-1711){\makebox(0,0)[lb]{\smash{\SetFigFont{8}{9.6}{\rmdefault}{\mddefault}{\updefault}$C_{q_1}$}}}
\put(4651,-2236){\makebox(0,0)[lb]{\smash{\SetFigFont{8}{9.6}{\rmdefault}{\mddefault}{\updefault}$C_{rev}$}}}
\put(3826,-961){\makebox(0,0)[lb]{\smash{\SetFigFont{9}{10.8}{\rmdefault}{\mddefault}{\updefault}$N_3$:  $\leftarrow p([b,f(a,b),X_1])$}}}
\put(3151,-1486){\makebox(0,0)[lb]{\smash{\SetFigFont{9}{10.8}{\rmdefault}{\mddefault}{\updefault}$N_4$:  $\leftarrow q([b,f(a,b),X_1],Z_2),reverse(Z_2,[],Z_3),p(Z_3)$}}}
\put(3151,-2011){\makebox(0,0)[lb]{\smash{\SetFigFont{9}{10.8}{\rmdefault}{\mddefault}{\updefault}$N_5$:  $\leftarrow reverse([X_2,f(b,f(a,b)),f(a,b),X_1],[],Z_3),p(Z_3)$}}}
\put(4726,-2911){\makebox(0,0)[lb]{\smash{\SetFigFont{8}{9.6}{\rmdefault}{\mddefault}{\updefault}$C_{p_1},C_{q_1},C_{rev}$}}}
\put(3226,-3286){\makebox(0,0)[lb]{\smash{\SetFigFont{9}{10.8}{\rmdefault}{\mddefault}{\updefault}$N_9$:  $\leftarrow p([X_2,f(b,f(a,b)),f(a,b),f(X_1,f(a,b)),X_3])$}}}
\put(4726,-3661){\makebox(0,0)[lb]{\smash{\SetFigFont{8}{9.6}{\rmdefault}{\mddefault}{\updefault}$C_{p_1},C_{q_1},C_{rev}$}}}
\put(2701,-4036){\makebox(0,0)[lb]{\smash{\SetFigFont{9}{10.8}{\rmdefault}{\mddefault}{\updefault}$*N_{12}$:  $\leftarrow p([X_3,f(X_1,f(a,b)),f(a,b),f(b,f(a,b)),f(X_2,f(b,f(a,b))),X_4])$}}}
\end{picture}

\caption{A generalized SLDNF-derivation for $P_4\cup \{\leftarrow Q_1\}$}\label{fig4-0}
\end{figure}

}
\end{example}

\begin{example}[\cite{Apt1}]
\label{eg6}
{\em
Consider the following well-known game program:
\begin{tabbing}
$\qquad$ \= $P_5:$ $\quad$ \= $win(X)\leftarrow move(X,Y). \neg win(Y)$. \`$C_{w_1}$\\
\>\> $move(a,b)\leftarrow$ $for$ $(a,b)\in \cal{G}$ where $\cal{G}$
is an acyclic finite graph.       \`$C_{m_1}$ 
\end{tabbing}
Assume the following two types of queries:
\begin{tabbing}
$\qquad$ \= $Q_1= win(a)$,\\
\> $Q_2= win(X)$.
\end{tabbing}

Since $\cal{G}$ is an acyclic finite graph, no
expanded variants occur in any generalized SLDNF-derivations. Therefore,
Algorithm \ref{alg1} will terminate for both $Q_1$ and $Q_2$ with
an answer {\em Yes}. That is, $P_5$ is terminating w.r.t.
$\{Q_1,Q_2\}$.
}
\end{example}

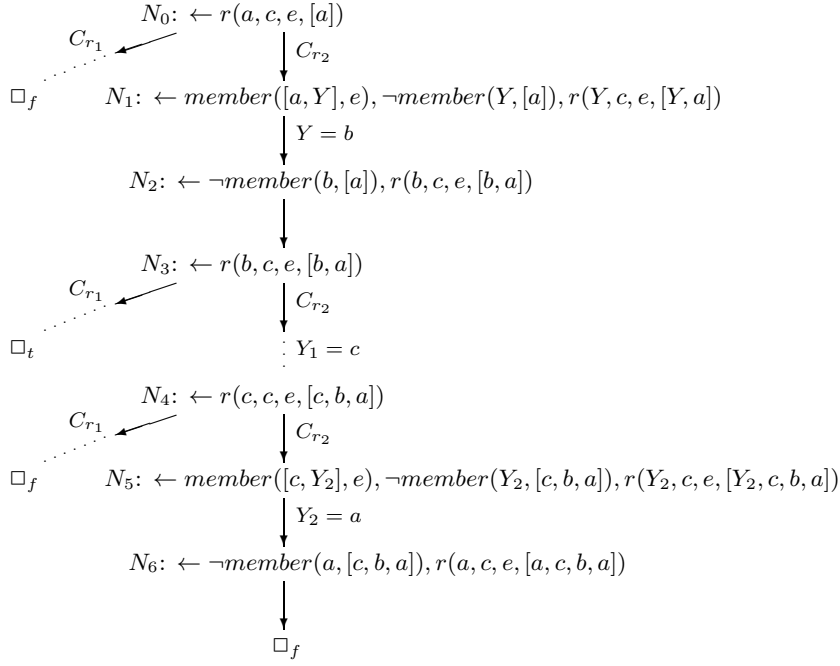
\begin{figure}[p]
$\qquad\quad$
\setlength{\unitlength}{3947sp}%
\begingroup\makeatletter\ifx\SetFigFont\undefined%
\gdef\SetFigFont#1#2#3#4#5{%
  \reset@font\fontsize{#1}{#2pt}%
  \fontfamily{#3}\fontseries{#4}\fontshape{#5}%
  \selectfont}%
\fi\endgroup%
\begin{picture}(1800,4080)(2851,-3361)
\thinlines
\put(3901,539){\vector(-3,-1){382.500}}
\multiput(3451,389)(-62.50000,-25.00000){7}{\makebox(1.6667,11.6667){\SetFigFont{5}{6}{\rmdefault}{\mddefault}{\updefault}.}}
\put(3901,-1036){\vector(-3,-1){382.500}}
\multiput(3451,-1186)(-62.50000,-25.00000){7}{\makebox(1.6667,11.6667){\SetFigFont{5}{6}{\rmdefault}{\mddefault}{\updefault}.}}
\multiput(4576,-1336)(0.00000,-75.00000){4}{\makebox(1.6667,11.6667){\SetFigFont{5}{6}{\rmdefault}{\mddefault}{\updefault}.}}
\put(3901,-1861){\vector(-3,-1){382.500}}
\multiput(3451,-2011)(-62.50000,-25.00000){7}{\makebox(1.6667,11.6667){\SetFigFont{5}{6}{\rmdefault}{\mddefault}{\updefault}.}}
\put(4576,539){\vector( 0,-1){300}}
\put(4576, 14){\vector( 0,-1){300}}
\put(4576,-511){\vector( 0,-1){300}}
\put(4576,-1036){\vector( 0,-1){300}}
\put(4576,-1861){\vector( 0,-1){300}}
\put(4576,-2386){\vector( 0,-1){300}}
\put(4576,-2911){\vector( 0,-1){300}}
\put(4651,-136){\makebox(0,0)[lb]{\smash{\SetFigFont{8}{9.6}{\rmdefault}{\mddefault}{\updefault}$Y=b$}}}
\put(4651,389){\makebox(0,0)[lb]{\smash{\SetFigFont{8}{9.6}{\rmdefault}{\mddefault}{\updefault}$C_{r_2}$}}}
\put(3451, 89){\makebox(0,0)[lb]{\smash{\SetFigFont{9}{10.8}{\rmdefault}{\mddefault}{\updefault}$N_1$:  $\leftarrow member([a,Y],e),\neg member(Y,[a]),r(Y,c,e,[Y,a])$ }}}
\put(3601,-436){\makebox(0,0)[lb]{\smash{\SetFigFont{9}{10.8}{\rmdefault}{\mddefault}{\updefault}$N_2$:  $\leftarrow \neg member(b,[a]),r(b,c,e,[b,a])$ }}}
\put(3676,614){\makebox(0,0)[lb]{\smash{\SetFigFont{9}{10.8}{\rmdefault}{\mddefault}{\updefault}$N_0$:  $\leftarrow r(a,c,e,[a])$}}}
\put(2851, 89){\makebox(0,0)[lb]{\smash{\SetFigFont{9}{10.8}{\rmdefault}{\mddefault}{\updefault}$\Box_f$}}}
\put(3226,464){\makebox(0,0)[lb]{\smash{\SetFigFont{8}{9.6}{\rmdefault}{\mddefault}{\updefault}$C_{r_1}$}}}
\put(3676,-961){\makebox(0,0)[lb]{\smash{\SetFigFont{9}{10.8}{\rmdefault}{\mddefault}{\updefault}$N_3$:  $\leftarrow r(b,c,e,[b,a])$ }}}
\put(2851,-1486){\makebox(0,0)[lb]{\smash{\SetFigFont{9}{10.8}{\rmdefault}{\mddefault}{\updefault}$\Box_t$}}}
\put(3226,-1111){\makebox(0,0)[lb]{\smash{\SetFigFont{8}{9.6}{\rmdefault}{\mddefault}{\updefault}$C_{r_1}$}}}
\put(4651,-1186){\makebox(0,0)[lb]{\smash{\SetFigFont{8}{9.6}{\rmdefault}{\mddefault}{\updefault}$C_{r_2}$}}}
\put(4651,-1486){\makebox(0,0)[lb]{\smash{\SetFigFont{8}{9.6}{\rmdefault}{\mddefault}{\updefault}$Y_1=c$}}}
\put(3676,-1786){\makebox(0,0)[lb]{\smash{\SetFigFont{9}{10.8}{\rmdefault}{\mddefault}{\updefault}$N_4$:  $\leftarrow r(c,c,e,[c,b,a])$ }}}
\put(2851,-2311){\makebox(0,0)[lb]{\smash{\SetFigFont{9}{10.8}{\rmdefault}{\mddefault}{\updefault}$\Box_f$}}}
\put(3226,-1936){\makebox(0,0)[lb]{\smash{\SetFigFont{8}{9.6}{\rmdefault}{\mddefault}{\updefault}$C_{r_1}$}}}
\put(4651,-2011){\makebox(0,0)[lb]{\smash{\SetFigFont{8}{9.6}{\rmdefault}{\mddefault}{\updefault}$C_{r_2}$}}}
\put(3451,-2311){\makebox(0,0)[lb]{\smash{\SetFigFont{9}{10.8}{\rmdefault}{\mddefault}{\updefault}$N_5$:  $\leftarrow member([c,Y_2],e),\neg member(Y_2,[c,b,a]),r(Y_2,c,e,[Y_2,c,b,a])$ }}}
\put(4651,-2536){\makebox(0,0)[lb]{\smash{\SetFigFont{8}{9.6}{\rmdefault}{\mddefault}{\updefault}$Y_2=a$}}}
\put(3601,-2836){\makebox(0,0)[lb]{\smash{\SetFigFont{9}{10.8}{\rmdefault}{\mddefault}{\updefault}$N_6$:  $\leftarrow \neg member(a,[c,b,a]),r(a,c,e,[a,c,b,a])$ }}}
\put(4501,-3361){\makebox(0,0)[lb]{\smash{\SetFigFont{9}{10.8}{\rmdefault}{\mddefault}{\updefault}$\Box_f$}}}
\end{picture}

\caption{$GT_{\leftarrow Q_1}$ for $P_6\cup \{\leftarrow Q_1\}$}\label{fig4}
\end{figure}

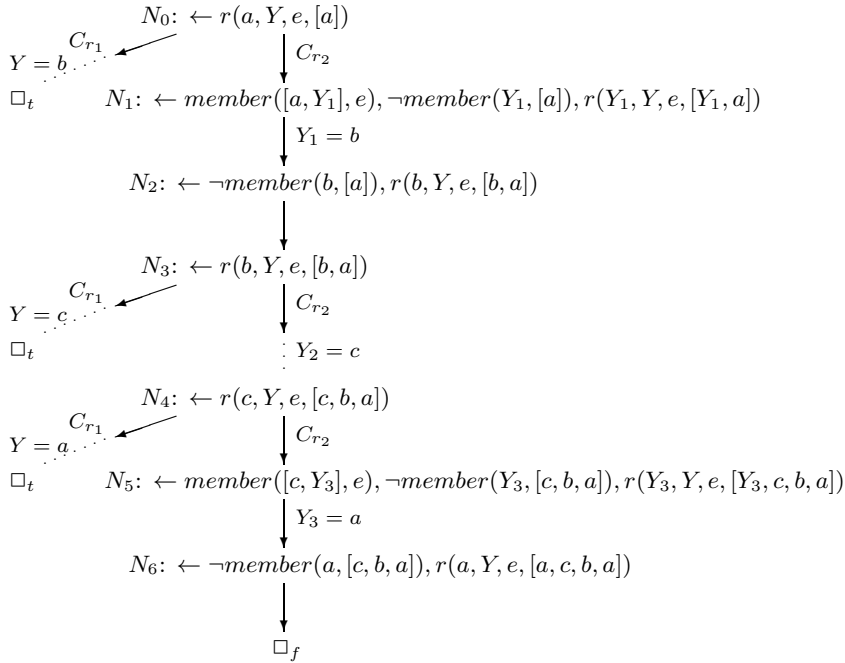
\begin{figure}[p]
$\qquad\quad$
\setlength{\unitlength}{3947sp}%
\begingroup\makeatletter\ifx\SetFigFont\undefined%
\gdef\SetFigFont#1#2#3#4#5{%
  \reset@font\fontsize{#1}{#2pt}%
  \fontfamily{#3}\fontseries{#4}\fontshape{#5}%
  \selectfont}%
\fi\endgroup%
\begin{picture}(1800,4080)(2851,-3361)
\thinlines
\put(3901,539){\vector(-3,-1){382.500}}
\multiput(3451,389)(-62.50000,-25.00000){7}{\makebox(1.6667,11.6667){\SetFigFont{5}{6}{\rmdefault}{\mddefault}{\updefault}.}}
\put(3901,-1036){\vector(-3,-1){382.500}}
\multiput(3451,-1186)(-62.50000,-25.00000){7}{\makebox(1.6667,11.6667){\SetFigFont{5}{6}{\rmdefault}{\mddefault}{\updefault}.}}
\multiput(4576,-1336)(0.00000,-75.00000){4}{\makebox(1.6667,11.6667){\SetFigFont{5}{6}{\rmdefault}{\mddefault}{\updefault}.}}
\put(3901,-1861){\vector(-3,-1){382.500}}
\multiput(3451,-2011)(-62.50000,-25.00000){7}{\makebox(1.6667,11.6667){\SetFigFont{5}{6}{\rmdefault}{\mddefault}{\updefault}.}}
\put(4576,539){\vector( 0,-1){300}}
\put(4576, 14){\vector( 0,-1){300}}
\put(4576,-511){\vector( 0,-1){300}}
\put(4576,-1036){\vector( 0,-1){300}}
\put(4576,-1861){\vector( 0,-1){300}}
\put(4576,-2386){\vector( 0,-1){300}}
\put(4576,-2911){\vector( 0,-1){300}}
\put(4651,-136){\makebox(0,0)[lb]{\smash{\SetFigFont{8}{9.6}{\rmdefault}{\mddefault}{\updefault}$Y_1=b$}}}
\put(4651,389){\makebox(0,0)[lb]{\smash{\SetFigFont{8}{9.6}{\rmdefault}{\mddefault}{\updefault}$C_{r_2}$}}}
\put(3451, 89){\makebox(0,0)[lb]{\smash{\SetFigFont{9}{10.8}{\rmdefault}{\mddefault}{\updefault}$N_1$:  $\leftarrow member([a,Y_1],e),\neg member(Y_1,[a]),r(Y_1,Y,e,[Y_1,a])$ }}}
\put(3601,-436){\makebox(0,0)[lb]{\smash{\SetFigFont{9}{10.8}{\rmdefault}{\mddefault}{\updefault}$N_2$:  $\leftarrow \neg member(b,[a]),r(b,Y,e,[b,a])$ }}}
\put(3676,614){\makebox(0,0)[lb]{\smash{\SetFigFont{9}{10.8}{\rmdefault}{\mddefault}{\updefault}$N_0$:  $\leftarrow r(a,Y,e,[a])$}}}
\put(2851, 89){\makebox(0,0)[lb]{\smash{\SetFigFont{9}{10.8}{\rmdefault}{\mddefault}{\updefault}$\Box_t$}}}
\put(3226,464){\makebox(0,0)[lb]{\smash{\SetFigFont{8}{9.6}{\rmdefault}{\mddefault}{\updefault}$C_{r_1}$}}}
\put(3676,-961){\makebox(0,0)[lb]{\smash{\SetFigFont{9}{10.8}{\rmdefault}{\mddefault}{\updefault}$N_3$:  $\leftarrow r(b,Y,e,[b,a])$ }}}
\put(2851,-1486){\makebox(0,0)[lb]{\smash{\SetFigFont{9}{10.8}{\rmdefault}{\mddefault}{\updefault}$\Box_t$}}}
\put(3226,-1111){\makebox(0,0)[lb]{\smash{\SetFigFont{8}{9.6}{\rmdefault}{\mddefault}{\updefault}$C_{r_1}$}}}
\put(4651,-1186){\makebox(0,0)[lb]{\smash{\SetFigFont{8}{9.6}{\rmdefault}{\mddefault}{\updefault}$C_{r_2}$}}}
\put(3676,-1786){\makebox(0,0)[lb]{\smash{\SetFigFont{9}{10.8}{\rmdefault}{\mddefault}{\updefault}$N_4$:  $\leftarrow r(c,Y,e,[c,b,a])$ }}}
\put(2851,-2311){\makebox(0,0)[lb]{\smash{\SetFigFont{9}{10.8}{\rmdefault}{\mddefault}{\updefault}$\Box_t$}}}
\put(3226,-1936){\makebox(0,0)[lb]{\smash{\SetFigFont{8}{9.6}{\rmdefault}{\mddefault}{\updefault}$C_{r_1}$}}}
\put(4651,-2011){\makebox(0,0)[lb]{\smash{\SetFigFont{8}{9.6}{\rmdefault}{\mddefault}{\updefault}$C_{r_2}$}}}
\put(3451,-2311){\makebox(0,0)[lb]{\smash{\SetFigFont{9}{10.8}{\rmdefault}{\mddefault}{\updefault}$N_5$:  $\leftarrow member([c,Y_3],e),\neg member(Y_3,[c,b,a]),r(Y_3,Y,e,[Y_3,c,b,a])$ }}}
\put(4651,-2536){\makebox(0,0)[lb]{\smash{\SetFigFont{8}{9.6}{\rmdefault}{\mddefault}{\updefault}$Y_3=a$}}}
\put(3601,-2836){\makebox(0,0)[lb]{\smash{\SetFigFont{9}{10.8}{\rmdefault}{\mddefault}{\updefault}$N_6$:  $\leftarrow \neg member(a,[c,b,a]),r(a,Y,e,[a,c,b,a])$ }}}
\put(4501,-3361){\makebox(0,0)[lb]{\smash{\SetFigFont{9}{10.8}{\rmdefault}{\mddefault}{\updefault}$\Box_f$}}}
\put(4651,-1486){\makebox(0,0)[lb]{\smash{\SetFigFont{8}{9.6}{\rmdefault}{\mddefault}{\updefault}$Y_2=c$}}}
\put(2851,314){\makebox(0,0)[lb]{\smash{\SetFigFont{8}{9.6}{\rmdefault}{\mddefault}{\updefault}$Y=b$}}}
\put(2851,-1261){\makebox(0,0)[lb]{\smash{\SetFigFont{8}{9.6}{\rmdefault}{\mddefault}{\updefault}$Y=c$}}}
\put(2851,-2086){\makebox(0,0)[lb]{\smash{\SetFigFont{8}{9.6}{\rmdefault}{\mddefault}{\updefault}$Y=a$}}}
\end{picture}

\caption{$GT_{\leftarrow Q_2}$ for $P_6\cup \{\leftarrow Q_2\}$}\label{fig5}
\end{figure}

\begin{figure}[htb]
$\qquad\quad$
\setlength{\unitlength}{3947sp}%
\begingroup\makeatletter\ifx\SetFigFont\undefined%
\gdef\SetFigFont#1#2#3#4#5{%
  \reset@font\fontsize{#1}{#2pt}%
  \fontfamily{#3}\fontseries{#4}\fontshape{#5}%
  \selectfont}%
\fi\endgroup%
\begin{picture}(4512,1936)(2026,-1217)
\thinlines
\put(3901,539){\vector(-3,-1){382.500}}
\multiput(3451,389)(-62.50000,-25.00000){7}{\makebox(1.6667,11.6667){\SetFigFont{5}{6}{\rmdefault}{\mddefault}{\updefault}.}}
\put(3676, 14){\vector(-3,-1){382.500}}
\multiput(3226,-136)(-62.50000,-25.00000){7}{\makebox(1.6667,11.6667){\SetFigFont{5}{6}{\rmdefault}{\mddefault}{\updefault}.}}
\put(5701, 14){\vector( 3,-1){382.500}}
\multiput(6151,-136)(62.50000,-25.00000){7}{\makebox(1.6667,11.6667){\SetFigFont{5}{6}{\rmdefault}{\mddefault}{\updefault}.}}
\put(2251, 89){\makebox(0,0)[lb]{\smash{\SetFigFont{9}{10.8}{\rmdefault}{\mddefault}{\updefault}Terminating}}}
\put(4201,-1186){\makebox(0,0)[lb]{\smash{\SetFigFont{9}{10.8}{\rmdefault}{\mddefault}{\updefault}Terminating}}}
\put(6001,-1186){\makebox(0,0)[lb]{\smash{\SetFigFont{9}{10.8}{\rmdefault}{\mddefault}{\updefault}Terminating}}}
\put(2476,-1186){\makebox(0,0)[lb]{\smash{\SetFigFont{9}{10.8}{\rmdefault}{\mddefault}{\updefault}Terminating}}}
\put(4576,539){\vector( 0,-1){300}}
\put(2851,-586){\vector( 0,-1){200}}
\put(4576,-586){\vector( 0,-1){200}}
\put(6376,-586){\vector( 0,-1){200}}
\put(4576, 14){\vector( 0,-1){200}}
\multiput(4576,-261)(0.00000,-50.00000){3}{\makebox(1.6667,11.6667){\SetFigFont{5}{6}{\rmdefault}{\mddefault}{\updefault}.}}
\multiput(2851,-886)(0.00000,-50.00000){3}{\makebox(1.6667,11.6667){\SetFigFont{5}{6}{\rmdefault}{\mddefault}{\updefault}.}}
\multiput(4576,-886)(0.00000,-50.00000){3}{\makebox(1.6667,11.6667){\SetFigFont{5}{6}{\rmdefault}{\mddefault}{\updefault}.}}
\multiput(6376,-886)(0.00000,-50.00000){3}{\makebox(1.6667,11.6667){\SetFigFont{5}{6}{\rmdefault}{\mddefault}{\updefault}.}}
\put(4651,-136){\makebox(0,0)[lb]{\smash{\SetFigFont{8}{9.6}{\rmdefault}{\mddefault}{\updefault}$X=b,Y_1=c$}}}
\put(4651,389){\makebox(0,0)[lb]{\smash{\SetFigFont{8}{9.6}{\rmdefault}{\mddefault}{\updefault}$C_{r_2}$}}}
\put(3451, 89){\makebox(0,0)[lb]{\smash{\SetFigFont{9}{10.8}{\rmdefault}{\mddefault}{\updefault}$N_1$:  $\leftarrow member([X,Y_1],e),\neg member(Y_1,[X]),r(Y_1,Y,e,[Y_1,X])$ }}}
\put(3676,614){\makebox(0,0)[lb]{\smash{\SetFigFont{9}{10.8}{\rmdefault}{\mddefault}{\updefault}$N_0$:  $\leftarrow r(X,Y,e,[X])$}}}
\put(3226,464){\makebox(0,0)[lb]{\smash{\SetFigFont{8}{9.6}{\rmdefault}{\mddefault}{\updefault}$C_{r_1}$}}}
\put(6526,-211){\makebox(0,0)[lb]{\smash{\SetFigFont{8}{9.6}{\rmdefault}{\mddefault}{\updefault}$X=c,Y_1=a$}}}
\put(3901,-511){\makebox(0,0)[lb]{\smash{\SetFigFont{9}{10.8}{\rmdefault}{\mddefault}{\updefault}$N_3$:  $\leftarrow r(c,Y,e,[c.b])$ }}}
\put(5701,-511){\makebox(0,0)[lb]{\smash{\SetFigFont{9}{10.8}{\rmdefault}{\mddefault}{\updefault}$N_4$:  $\leftarrow r(a,Y,e,[a,c])$ }}}
\put(2176,-511){\makebox(0,0)[lb]{\smash{\SetFigFont{9}{10.8}{\rmdefault}{\mddefault}{\updefault}$N_2$:  $\leftarrow r(b,Y,e,[b,a])$ }}}
\put(2026,-211){\makebox(0,0)[lb]{\smash{\SetFigFont{8}{9.6}{\rmdefault}{\mddefault}{\updefault}$X=a,Y_1=b$}}}
\end{picture}

\caption{$GT_{\leftarrow Q_3}$ for $P_6\cup \{\leftarrow Q_3\}$}\label{fig6}
\end{figure}
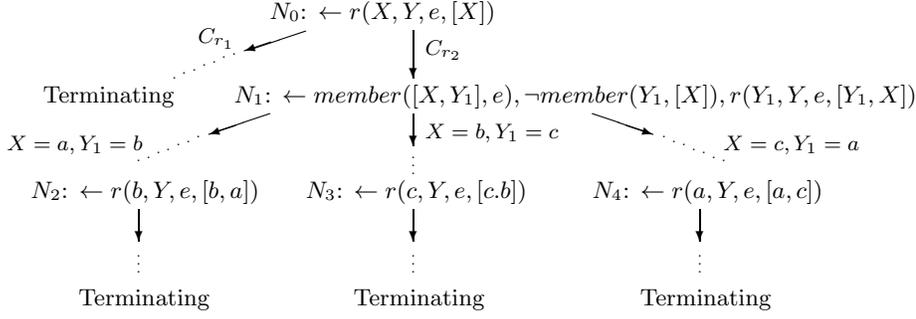

\begin{example}[\cite{Apt1}]
\label{eg7}
{\em
The following general logic program is used to compute the transitive 
closure of a graph.
\begin{tabbing}
$\quad$\= $P_6:$$\quad$\= $r(X,Y,E,V)\leftarrow member([X,Y],E)$. \`$C_{r_1}$\\
\>\> $r(X,Z,E,V)\leftarrow member([X,Y],E), \neg member(Y,V),
      r(Y,Z,E,[Y|V])$. \`$C_{r_2}$\\
\>\> $member(X,[X|T])$.      \`$C_{m_1}$\\
\>\> $member(X,[Y|T]) \leftarrow member(X,T)$.       \`$C_{m_1}$ 
\end{tabbing}
Queries over this program are of the form $r(X,Y,e,[X])$ where $X$, $Y$ are 
nodes and $e$ is a graph specified by a finite list of its edges denoted by $[Node,Node]$.
Such a query is supposed to succeed when $[X,Y]$ is in the transitive
closure of $e$. The last argument of $r(X,Y,e,[X])$ acts as an accumulator
in which a list of nodes is maintained which should not be reused when looking
for a path connecting $X$ with $Y$ in $e$ (to keep the search path acyclic).
As an example, let $e=\{[[a,b],[b,c],[c,a]]\}$. We consider the following
three types of queries:
\begin{tabbing}
$\qquad$ \= $Q_1= r(a,c,e,[a])$,\\
\> $Q_2= r(a,Y,e,[a])$,\\
\> $Q_3= r(X,Y,e,[X])$.
\end{tabbing}
The generalized SLDNF-trees $GT_{\leftarrow Q_1}$,
$GT_{\leftarrow Q_2}$, and $GT_{\leftarrow Q_3}$ are
depicted in Figures \ref{fig4}, \ref{fig5} and \ref{fig6}, respectively.
Since there is no expanded variant in any generalized SLDNF-derivations,
Algorithm \ref{alg1} will return {\em Yes} when executing 
$Test(P_6,\{Q_1,Q_2,Q_3\},R)$. That is, $P_6$
is terminating w.r.t. these three types of queries.

}
\end{example}

It is interesting to observe that for each of the above 
logic programs, $P_1-P_6$, it is terminating if and only if applying
Algorithm \ref{alg1} to it yields an answer {\em Yes}. In fact, this 
is true for all representative logic programs we have currently collected in the 
literature. However, due to the undecidability of the termination problem, 
it is unavoidable that there exist cases in which
Algorithm \ref{alg1} returns a generalized SLDNF-derivation $D$, but $P$ is 
a terminating logic program. We have created such a rarely used program.

\begin{example}
\label{eg8}
{\em
Consider the following logic program and top goal,
where the function $size(Z)$ returns the number of elements
in the list $Z$ (e.g. $size([a,b])=2$).
\begin{tabbing}
$\qquad$ \= $P_7:$ $\quad$ \= $p([X|Y],N)\leftarrow size([X|Y])<N,
        p([X,X|Y],N)$. \`$C_{p_1}$\\
\>          $G_0:$ \> $\leftarrow p([a],100).$
\end{tabbing}

The generalized SLDNF-tree $GT_{G_0}$ for
$P_7\cup \{G_0\}$ is shown in Figure \ref{fig7}.
It is easy to see that for any $i\geq 0$, the subgoal $L_{2*i}^1$ at $N_{2*i}$
is an ancestor subgoal of the subgoal $L_{2*(i+1)}^1$ at $N_{2*(i+1)}$, while
$L_{2*(i+1)}^1$ is an expanded variant of $L_{2*i}^1$ with 
$|L_{2*(i+1)}^1|>|L_{2*i}^1|$. 
$P_7$ is terminating w.r.t. the query $p([a],100).$ However,
applying Algorithm \ref{alg1} (with $d=2$) will return a generalized SLDNF-derivation
$D$ that is the segment between $N_0$ and $N_4$ in Figure \ref{fig7}.
Apparently, in order for Algorithm \ref{alg1} to return {\em Yes}
the depth bound $d$ should not be less than 100.
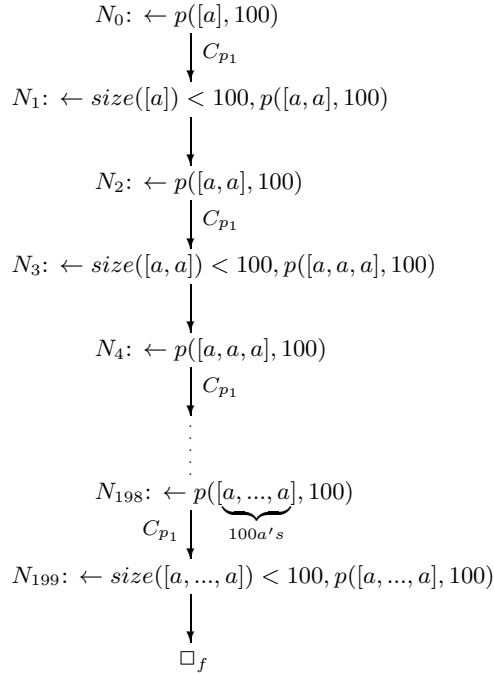
\begin{figure}[htb]
$\qquad\qquad\qquad$
\setlength{\unitlength}{3947sp}%
\begingroup\makeatletter\ifx\SetFigFont\undefined%
\gdef\SetFigFont#1#2#3#4#5{%
  \reset@font\fontsize{#1}{#2pt}%
  \fontfamily{#3}\fontseries{#4}\fontshape{#5}%
  \selectfont}%
\fi\endgroup%
\begin{picture}(1200,4146)(3451,-3436)
\thinlines
\multiput(4576,-1936)(0.00000,-75.00000){5}{\makebox(1.6667,11.6667){\SetFigFont{5}{6}{\rmdefault}{\mddefault}{\updefault}.}}
\put(4576,539){\vector( 0,-1){300}}
\put(4576,-1036){\vector( 0,-1){300}}
\put(4576,-1561){\vector( 0,-1){300}}
\put(4576, 14){\vector( 0,-1){300}}
\put(4576,-511){\vector( 0,-1){300}}
\put(4576,-2461){\vector( 0,-1){300}}
\put(4576,-2986){\vector( 0,-1){300}}
\put(3976,614){\makebox(0,0)[lb]{\smash{\SetFigFont{9}{10.8}{\rmdefault}{\mddefault}{\updefault}$N_0$:  $\leftarrow p([a],100)$}}}
\put(4651,389){\makebox(0,0)[lb]{\smash{\SetFigFont{8}{9.6}{\rmdefault}{\mddefault}{\updefault}$C_{p_1}$}}}
\put(3451,-961){\makebox(0,0)[lb]{\smash{\SetFigFont{9}{10.8}{\rmdefault}{\mddefault}{\updefault}$N_3$:  $\leftarrow size([a,a])<100,p([a,a,a],100)$}}}
\put(4651,-1711){\makebox(0,0)[lb]{\smash{\SetFigFont{8}{9.6}{\rmdefault}{\mddefault}{\updefault}$C_{p_1}$}}}
\put(3976,-1486){\makebox(0,0)[lb]{\smash{\SetFigFont{9}{10.8}{\rmdefault}{\mddefault}{\updefault}$N_4$:  $\leftarrow p([a,a,a],100)$}}}
\put(3451, 89){\makebox(0,0)[lb]{\smash{\SetFigFont{9}{10.8}{\rmdefault}{\mddefault}{\updefault}$N_1$:  $\leftarrow size([a])<100,p([a,a],100)$}}}
\put(4651,-661){\makebox(0,0)[lb]{\smash{\SetFigFont{8}{9.6}{\rmdefault}{\mddefault}{\updefault}$C_{p_1}$}}}
\put(3976,-436){\makebox(0,0)[lb]{\smash{\SetFigFont{9}{10.8}{\rmdefault}{\mddefault}{\updefault}$N_2$:  $\leftarrow p([a,a],100)$}}}
\put(3976,-2386){\makebox(0,0)[lb]{\smash{\SetFigFont{9}{10.8}{\rmdefault}{\mddefault}{\updefault}$N_{198}$:  $\leftarrow p([\underbrace{a,...,a}_{100 a's}],100)$}}}
\put(4501,-3436){\makebox(0,0)[lb]{\smash{\SetFigFont{9}{10.8}{\rmdefault}{\mddefault}{\updefault}$\Box_f$}}}
\put(3451,-2911){\makebox(0,0)[lb]{\smash{\SetFigFont{9}{10.8}{\rmdefault}{\mddefault}{\updefault}$N_{199}$:  $\leftarrow size([a,...,a])<100,p([a,...,a],100)$}}}
\put(4276,-2611){\makebox(0,0)[lb]{\smash{\SetFigFont{8}{9.6}{\rmdefault}{\mddefault}{\updefault}$C_{p_1}$}}}
\end{picture}

\caption{$GT_{\leftarrow p([a],100)}$ for 
$P_7\cup \{\leftarrow p([a],100)\}$}\label{fig7}
\end{figure}

}
\end{example}

\section{Related Work}
\label{related-work}
Our work is related to both termination analysis and loop checking.

\subsection{Work on Termination Analysis}
Concerning termination analysis, we refer the reader to the survey of
Decorte, De Schreye and Vandecasteele \cite{DD93,DDV99} 
for a comprehensive bibliography. 

There are two essential differences between existing termination
analysis techniques and ours. The first difference is that
theirs are static approaches, whereas ours is a dynamic one.
Static approaches only make use of the syntactic structure of
the source code of a logic program to establish some well-founded
conditions/constraints that, when satisfied, yield a termination 
proof. Since non-termination is caused by an infinite generalized
SLDNF-derivation, which contains some essential dynamic characteristics
(such as expanded variants and the repeated application of
some clauses) that are hard to capture in a static way,
static approaches appear to be less precise than a dynamic one.
For example, it is difficult to apply a static approach to prove
the termination of program $P_2$ in Example \ref{eg2} with respect to
a query pattern $p$. Moreover, although some static approaches
(e.g., see \cite{DV95,Pl90a,UVG88,VD91}) are 
automatizable, searching for an appropriate level mapping
or computing some interargument relations could be 
very complex. For our approach, the major work is to identify 
expanded variants, which is easy to complete.

The second difference is that existing methods are suitable for 
termination analysis with respect to query patterns, whereas ours
is for termination analysis with respect to concrete queries. The
advantage of using query patterns is that if a logic program $P$
is shown to be terminating with respect to a query pattern $Q$, it is
terminating with respect to all instances of $Q$ that could be an infinite
set of concrete queries. However, if $P$ is shown to be not terminating
with respect to $Q$, which usually means that $P$ is terminating with respect to
some instances of $Q$ but is not with respect to the others, we cannot
apply existing termination analysis methods to make such a further
distinction. In contrast, our method can make termination analysis for
each single concrete query posed by the user and provide 
explanations about how non-termination happens. This
turns out to be very useful in real programming practices. Observe that
in developing a software in any computer languages we always apply some typical
cases (i.e. concrete parameters as inputs) to test for the correctness
or termination of the underlying programs, with an assumption that
if the software works well with these typical cases, it would work well
with all cases of practical interests.

From the above discussion, it is easy to see that our method plays
a complementary role with respect to existing termination analysis
approaches (i.e. static versus dynamic and query patterns versus concrete
queries).

\subsection{Work on Loop Checking}
Loop checking is a run-time approach towards termination. It
locates nodes at which SLD-derivations step into
a loop and prunes them from SLD-trees. Informally, an SLD-derivation
\begin{center}
$N_0:G_0 \Rightarrow N_1:G_1 \Rightarrow ...\Rightarrow N_i:G_i \Rightarrow ...
\Rightarrow N_k:G_k \Rightarrow ...$
\end{center}
is said to step into a loop at a node $N_k:G_k$ if there is a
node $N_i:G_i$ ($0 \leq i < k$) in the
derivation such that $G_i$ and $G_k$ are {\em sufficiently
similar}. Many mechanisms related to loop checking 
have been presented in the literature
(e.g. see \cite{BAK91,BDM92,Cov85,FPS95,MDB92,MD96,
Sa93,shen97,shen001,Sk97,SGG86,VG87,VL89}).
We mention here a few representative ones.

Bol, Apt and Klop \cite{BAK91} introduced the {\em Equality check} and
the {\em Subsumption check}. These loop checks can detect loops of the form  
\begin{center}
$N_0:G_0 \Rightarrow N_1:G_1 \Rightarrow ...\Rightarrow N_i:G_i \Rightarrow ...
\Rightarrow N_k:G_k$
\end{center}
where either $G_k$ is a variant or an instance of $G_i$ (for the Equality check), 
i.e. $G_k=G_i\theta$ under a substitution $\theta$, or $G_i$ is included in
$G_k$ under a substitution $\theta$ (for the Subsumption check), i.e.
$G_k\supseteq G_i\theta$. However, they cannot handle infinite SLD-derivations
of the form
\begin{center}
$N_0:p(X) \Rightarrow N_1:p(f(X)) \Rightarrow ...\Rightarrow 
N_i:p(f(...f(X)...)) \Rightarrow ...$
\end{center}

Sahlin \cite{Sa90,Sa93} introduced the {\em OS-check} (see also \cite{Bol93}).
It determines infinite loops based on two parameters: a depth bound 
$d$ and a size function {\em size}. Informally, OS-check says that an 
SLD-derivation may go into an infinite loop if it generates an 
oversized subgoal. A subgoal $A$ is said to be {\em oversized} if 
it has $d$ ancestor subgoals in the SLD-derivation that have the same 
predicate symbol as $A$ and whose size is smaller than or equal to $A$.

Bruynooghe, De schreye and Martens \cite{BDM92,MD96,MDB92} presented a framework
for partial deduction with finite unfolding that, when applied to loop checking,
is very similar to OS-check. That is, it mainly relies on term sizes of
(selected) subgoals and a depth bound. See \cite{Bol93,MD96} for a detailed
comparison of these works.

OS-check (similarly the method of Bruynooghe, De schreye and Martens) 
is {\em complete} in the sense that
it cuts all infinite loops. However, because it merely takes the
number of repeated predicate symbols and the size of subgoals as 
its decision parameters, without referring to the informative internal
structure of the subgoals, the underlying decision is fairly unreliable;
i.e. many non-loop derivations may be pruned unless the depth bound 
$d$ is set sufficiently large. 

Using expanded variants, in \cite{shen001} we proposed a series of 
loop checks, called {\em VAF-checks}
(for {\it V}ariant {\it A}toms loop checks 
for logic programs with {\it F}unctions). These loop
checks are complete and much more reliable than OS-check. However, they 
cannot deal with infinite recursions through negation like that in Figure \ref{fig1}.

The work of the current paper can partly be viewed as an extension of \cite{shen001}
from identifying infinite SLD-derivations to identifying 
infinite generalized SLDNF-derivations. It is worth noting that termination analysis
is merely concerned with the characterization and identification of infinite
derivations, but loop checking is also concerned about how to prune infinite
derivations. The latter work heavily relies on the semantics of a logic program,
especially when an infinite recursion through negation occurs. 
Bol \cite{Bol93} discussed loop checking for locally stratified logic programs
under the perfect model semantics \cite{Prz88}.

\section{Conclusions}
We have presented a method of verifying termination of
general logic programs with respect to concrete queries.
A necessary and sufficient condition is
established and an algorithm for automatic testing
is developed. Unlike existing termination analysis 
approaches, our method does not need to 
search for a model or a level mapping, nor does it need to compute
an interargument relation based on additional mode or type information. Instead, it 
detects infinite derivations by directly evaluating the set of
queries of interest. As a result, some key dynamic features of a
logic program can be extracted and employed to predict its termination. 
Such idea partly comes from loop checking. Therefore,
the work of this paper bridges termination analysis with loop checking,
the two problems which have been studied separately in the past despite their close 
technical relation with each other. 

It is worth mentioning that the practical purpose of termination 
analysis is to assist users to write terminating programs. Our method
exactly serves for this purpose. When 
Algorithm \ref{alg1} outputs {\em Yes}, the logic program is terminating;
otherwise it provides users with a generalized SLDNF-derivation 
of the form as shown in Figures \ref{fig3} or \ref{fig4-0}.
Such a derivation may most likely lead to an infinite derivation,
thus users can improve their programs following the informative guidance. (In this
sense, our method is quite like a {\em spelling} mechanism used
in a word processing system, which always indicates most likely 
incorrect spellings.)

Due to the undecidability of the termination problem,
there exist cases in which a logic program is terminating but 
Algorithm \ref{alg1} would not say {\em Yes} unless the depth bound
$d$ is set sufficiently large (see Example \ref{eg8}).
Although $d=2$ works well for a vast majority of logic programs
(see Examples \ref{eg3} - \ref{eg7}), how to choose the depth bound
in a general case then presents an interesting open problem.

Tabled logic programming is receiving increasing attention 
in the community of logic programming
(e.g. see \cite{BD98,chen96,shen99,shen002,TS86,VL89,war92}). 
Verbaeten, De Schreye and K. Sagonas \cite{VDK001} 
recently exploited termination proofs for positive logic programs
with tabling. For future research, we are going to extend the work of 
the current paper to deal with general logic programs with tabling.

\section*{Acknowledgements}
We would like to thank Danny De Schreye for his constructive comments on our work
and valuable suggestions for its improvement.

\end{document}